\documentclass{article}

\usepackage{microtype}
\usepackage{graphicx}
\usepackage{subfigure}
\usepackage{booktabs} 

\usepackage{hyperref}

\usepackage[accepted]{icml2025}

\usepackage{amsmath}
\usepackage{amssymb}
\usepackage{mathtools}
\usepackage{amsthm}

\usepackage[capitalize,noabbrev]{cleveref}

\theoremstyle{plain}
\newtheorem{theorem}{Theorem}[section]

\theoremstyle{definition}
\newtheorem{definition}[theorem]{Definition}

\theoremstyle{remark}

\usepackage[textsize=tiny]{todonotes}

\usepackage{colortbl}
\usepackage{wrapfig}
\usepackage[utf8]{inputenc} 
\usepackage[T1]{fontenc}    
\usepackage{url}            
\usepackage{booktabs}       
\usepackage{amsfonts}       
\usepackage{nicefrac}       
\usepackage{microtype}      
\usepackage{xcolor}         

\usepackage{soul}
\usepackage{url}
\usepackage[utf8]{inputenc}
\usepackage[small]{caption}
\usepackage{graphicx}
\usepackage{amsmath}
\usepackage{amsthm}
\usepackage{booktabs}
\usepackage{algorithm}
\usepackage{algorithmic}
\usepackage[switch]{lineno}

\usepackage{xspace}

\usepackage{newfloat}
\usepackage{listings}
\usepackage{comment}

\usepackage{caption}
\usepackage{subcaption}
\usepackage{enumitem}
\usepackage{lineno}
\usepackage{amsmath,amssymb,amsfonts}
\usepackage{algorithm}
\usepackage{algorithmic}
\usepackage{graphicx}
\usepackage{textcomp}
\usepackage{xcolor}
\usepackage{amsthm}
\usepackage{url}
\usepackage{multirow}
\usepackage{multicol}
\usepackage{color}
\usepackage{bm}
\usepackage{mdframed}

\usepackage{pifont}
\usepackage{array}
\newcolumntype{L}[1]{>{\raggedright\let\newline\\\arraybackslash\hspace{0pt}}m{#1}}
\newcolumntype{C}[1]{>{\centering\let\newline  \\\arraybackslash\hspace{0pt}}m{#1}}
\newcolumntype{R}[1]{>{\raggedleft\let\newline \\\arraybackslash\hspace{0pt}}m{#1}}
\renewcommand{\algorithmicrequire}{\textbf{Input:}}
\renewcommand{\algorithmicensure}{\textbf{Output:}}

\newcommand{\methodname}{\texttt{RBP}}

\newmdenv[
    linewidth=2pt,       
    roundcorner=10pt,    
    linecolor=gray!90,     
    backgroundcolor=gray!05, 
    skipabove=5pt,      
    skipbelow=5pt       
]{custombox}

\icmltitlerunning{Quantifying Explanatory Inversion in Post-Hoc Model Explanations}

\begin{document}

\twocolumn[
\icmltitle{\textbf{Are We Merely Justifying Results ex Post Facto?\\ Quantifying Explanatory Inversion in Post-Hoc Model Explanations
}}



\icmlsetsymbol{equal}{*}

\begin{icmlauthorlist}
\icmlauthor{Zhen Tan}{yyy}
\icmlauthor{Song Wang}{zzz}
\icmlauthor{Yifan Li}{xxx}
\icmlauthor{Yu Kong}{xxx}
\icmlauthor{Jundong Li}{zzz}
\icmlauthor{Tianlong Chen}{www}
\icmlauthor{Huan Liu}{yyy}

\end{icmlauthorlist}

\icmlaffiliation{yyy}{Arizona State University}
\icmlaffiliation{zzz}{University of Virginia}
\icmlaffiliation{xxx}{Michigan State University}
\icmlaffiliation{www}{University of North Carolina at Chapel Hill
}



\icmlkeywords{Machine Learning, ICML}

\vskip 0.3in
]



\printAffiliationsAndNotice{}  

\begin{abstract}
Post-hoc explanation methods provide interpretation
by attributing predictions to input features. Natural explanations are expected to interpret how the inputs lead to the predictions.
Thus, a {fundamental question} arises: \emph{Do these explanations unintentionally reverse the natural relationship between inputs and outputs?} {Specifically, are the explanations rationalizing predictions from the output rather than reflecting the true decision process?} To investigate such \emph{{explanatory inversion}}, we propose \emph{Inversion Quantification} (\emph{IQ}), a framework that quantifies the degree to which explanations rely on outputs and deviate from faithful input-output relationships.
Using the framework, we demonstrate on synthetic datasets that widely used methods such as LIME and SHAP are prone to such inversion, particularly in the presence of spurious correlations, across \textbf{tabular}, \textbf{image}, and \textbf{text} domains. 
Finally, we propose \emph{Reproduce-by-Poking} (\emph{\methodname}), a simple and {model-agnostic} enhancement to post-hoc explanation methods that integrates forward perturbation checks. We further show that under the IQ framework, \methodname~theoretically guarantees the mitigation of explanatory inversion.
Empirically, for example, on the synthesized data, \methodname~can reduce the inversion by $1.8\%$ on average across iconic post-hoc explanation approaches and domains. 

\end{abstract}

\section{Introduction}
\label{sec:introduction}

\begin{figure}[t]
    \centering
    \includegraphics[width=1\linewidth]{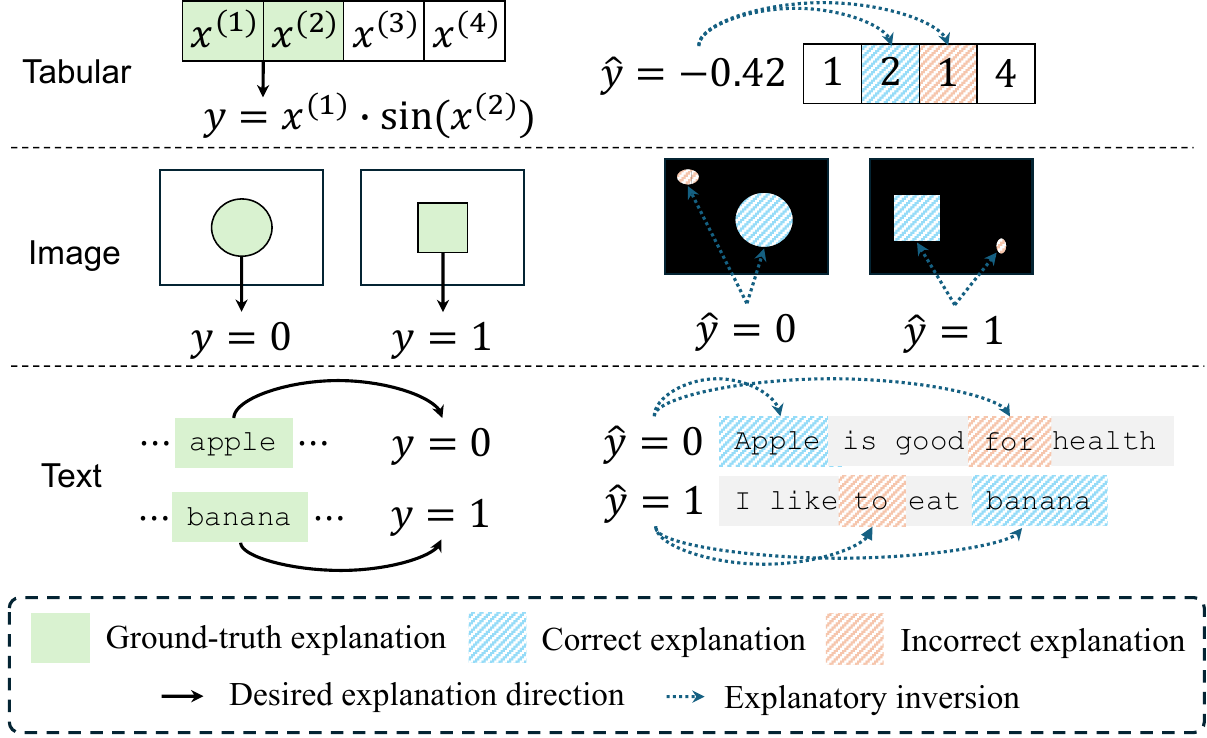}
    \caption{Illustration of post-hoc explanation methods and the potential of explanatory inversion. 
For tabular data (first row), ground-truth explanations attribute the output \(y = x^{(1)} \cdot sin(x^{(2)})\) to the contributions of features \(x^{(1)}\) and \(x^{(2)}\), but explanatory inversion misattributes \(x^{(2)}\) and \(x^{(3)}\). 
For image data (second row), explanations should focus on the correct object, but explanatory inversion leads to incorrect focus regions. 
For text data (third row), ground-truth explanations link keywords to labels, but explanatory inversion results in misaligned or irrelevant attributions. 
}\label{fig:intro-post-hoc}
    \vspace{-7mm}
\end{figure}

Post-hoc explanation methods have become essential tools for interpreting the predictions of complex machine learning (ML) models, particularly in high-stakes applications such as healthcare~\cite{turbe2023evaluation}, finance~\cite{de2024explainable}, and policy-making~\cite{heesen2024model}. By providing insights into which input features are most influential for a given output, methods such as SHAP~\cite{lundberg2017unified}, LIME~\cite{ribeiro2016should}, and Integrated Gradients~\cite{sundararajan2017axiomatic} aim to increase the transparency of predictive models.
Post-hoc explanation methods provide interpretation by attributing predictions to input features, while natural explanations are expected to interpret how
the inputs result in the predictions.
Therefore, a critical question remains unresolved:
\begin{custombox}
    \hypertarget{RQ}{\textbf{\textit{RQ}}:} \textit{Do post-hoc explanations faithfully represent the model’s decision-making process, without inadvertently rationalizing predictions from the output?}
\end{custombox}
\vspace{-1mm}
This potential reversal of reasoning, which we term \emph{explanatory inversion}, undermines the reliability of explanations and poses significant challenges to the broader adoption of AI systems in sensitive domains.

Explanatory inversion arises when a post-hoc explanation method over-relies on the model’s output in generating attributions, rather than accurately reflecting the relationship between inputs and predictions. For instance, a method may highlight features that appear important solely because they correlate with the model’s output, rather than because they genuinely influence the prediction. Understanding and quantifying such inversion is essential for identifying the limitations of existing explanation techniques and for developing more robust alternatives.
To investigate and mitigate explanatory inversion, we make these contributions:
\vspace{-3mm}
\begin{itemize}[leftmargin=*,itemsep=1.5pt]
    \item \textbf{Formal Definition and Framework:} We formally define explanatory inversion and introduce the \emph{Inversion Quatification} (\emph{IQ}), a novel framework that quantifies the degree to which explanations rely on outputs and deviate from faithful input-output relationships. IQ evaluates explanations along two key dimensions: \emph{reliance on outputs}, measuring the correlation between attributions and model predictions, and \emph{faithfulness}, assessing alignment with perturbations of input features. IQ is applicable to real-world data without ground-truth input importance.
    \vspace{-1mm}
    \item \textbf{Empirical Validation:} Using synthetic datasets across tabular, image, and text domains, we systematically verify the \textbf{presence} of explanatory inversion in widely used methods such as LIME and SHAP. Our experiments reveal that these methods are particularly vulnerable to spurious correlations, leading to significant explanatory inversion.
    \vspace{-1mm}
    \item \textbf{Mitigation via Reproduce-by-Poking:} We propose \emph{Reproduce-by-Poking} (\emph{\methodname}), a simple and model-agnostic enhancement to post-hoc explanation methods that incorporates forward perturbation checks. Under the IQ framework, we demonstrate that \methodname~provably mitigates explanatory inversion. For example, on synthesized data, \methodname~reduces inversion by \textbf{1.8\%} on average across iconic post-hoc explanation approaches and domains.
\end{itemize}

\vspace{-2mm}
\section{Related Work}\label{sec:related}
\vspace{-1mm}

\noindent\textbf{Post-hoc Explanation Methods.} Post-hoc techniques have become indispensable for interpreting predictions of complex models. SHAP~\cite{lundberg2017unified} leverages game-theoretic principles to ensure attributions satisfy desirable properties such as additivity and consistency. LIME~\cite{ribeiro2016should} approximates local model behavior using interpretable surrogate models, while Integrated Gradients~\cite{sundararajan2017axiomatic} integrates gradients along a path from a baseline to the input, ensuring completeness and sensitivity. Other methods include Occlusion~\cite{zeiler2014visualizing}, which computes feature importance by masking input regions, and SmoothGrad~\cite{smilkov2017smoothgrad}, which reduces noise in saliency maps via input perturbations. Post-hoc explanation techniques are widely adopted but are shown to rely heavily on model outputs, based on our results, raising concerns about their susceptibility to explanatory inversion~\cite{rodis2024multimodal}.
Recent advancements in post-hoc explanation methods have led to the development of more techniques to handle specific tasks~\cite{turbe2023evaluation,leemann2023post,parekh2022listen}. 
%
%
We give a more detailed description in Appendix~\ref{app:related}.

\vspace{-1mm}
\noindent\textbf{Evaluating Explanation Reliability.} Ensuring the reliability of explanations is a critical area of research. \citet{adebayo2018sanity} demonstrated that many explanation methods fail sanity checks, producing identical attributions even when models are randomized. \citet{hooker2019benchmark} introduced \emph{removal-based benchmarks} to assess explanation faithfulness by measuring performance degradation when features are removed. \citet{mohseni2021multidisciplinary} reviewed challenges in explaining deep models with domain-specific benchmarks. Focusing on robustness, they do not discuss the potential of explanation's dependence on outputs.

\vspace{-1mm}
\noindent\textbf{Bias and Vulnerabilities in Explanations.} Explanations are often sensitive to spurious correlations, adversarial perturbations, or data distribution shifts~\cite{alvarez2018robustness, slack2020fooling}. \citet{dimanov2020you} analyzed fairness in post-hoc explanations, showing that feature importance methods can perpetuate biases present in the data. \citet{wang2020explanations} examined how explanations can be manipulated to justify specific outcomes, even when they misalign with the model’s behavior. Our work extends these studies by introducing the Inversion Score (IS), which explicitly quantifies reliance on outputs and explanation faithfulness. IS remains applicable in the absence of ground-truth input importance, broadening its utility.

\vspace{-2mm}
\section{Preliminaries: Post-hoc Explanation}\label{sec:pre}
\vspace{-1mm}

In this section, we provide an overview of the explanation methods\footnote{These four methods were chosen because they are among the most widely used and are applicable across multiple models and domains. 
We acknowledge that other domain-specific methods, such as GradCAM \cite{selvaraju2017grad} for image data and attention scores for transformers, could also be employed for interpretation. However, we focus on general methods to maintain consistency across domains and leave the exploration of those domain-specific methods as future work. See Appendix~\ref{app:related}.} used across the tabular, image, and text domains, including SHAP, LIME, Integrated Gradients (IG), and Occlusion. Each method is formally defined as follows:

\vspace{-1mm}
\noindent$\rhd$~\textbf{SHAP} (SHapley Additive exPlanations, \citet{lundberg2017unified}) is a method utilizing cooperative game theory that aims to fairly distribute the model’s output among its input features based on their individual contributions. It attributes a model's output $f(x)$ for an input $x$ to features based on the Shapley value. The attribution for feature $i$ is defined as:
\begin{equation}\small
\phi_i = \sum_{S \subseteq \{1, \dots, d\} \setminus \{i\}} \frac{|S|!(d - |S| - 1)!}{d!} \left[ f(x_{S \cup \{i\}}) - f(x_S) \right],
\end{equation}
where $x_S$ is the input restricted to the subset $S$ and $d$ is the total number of features. SHAP is applicable across tabular, image, and text data due to its versatility in providing feature-level attributions.


\noindent$\rhd$~\textbf{LIME} (Local Interpretable Model-Agnostic Explanations, \citet{ribeiro2016should}) explains a model's prediction by locally approximating it with a simpler interpretable model, typically a linear regression. For an input $x$, LIME constructs a surrogate model $g \in G$, where $g$ is a simple interpretable function (e.g., linear regression), and $G$ is the space of such interpretable models. The weights of $g$ correspond to the attributions of features. The explanation is generated by perturbing the input, evaluating the model $f$, and minimizing the following objective:
\vspace{-1mm}
\begin{equation}
\arg\min_{g \in G} \sum_{x' \in \mathcal{N}(x)} \pi_x(x') \left( f(x') - g(x') \right)^2 + \Omega(g),
\vspace{-1mm}
\end{equation}
where $\mathcal{N}(x)$ is the set of perturbed samples, $\pi_x(x')$ is a proximity function measuring the similarity between $x'$ and $x$, and $\Omega(g)$ is a complexity penalty for the interpretable model $g$. LIME is model-agnostic and widely used for local explanations in multiple modalities, as it provides feature-level attributions based on the weights of the surrogate model.

\vspace{-1mm}
\noindent$\rhd$~\textbf{IG} (Integrated Gradients, \citet{sundararajan2017axiomatic}) attributes the prediction difference between an input \( x \) and a baseline \( x' \) to each feature. The baseline \( x' \) is a reference input that represents the absence of a meaningful signal, such as a black image for vision tasks or an empty/masked input for text. The attribution for feature \( i \) is computed as:
\vspace{-1mm}
\begin{equation}
\text{IG}_i = (x_i - x_i') \int_{\alpha=0}^1 \frac{\partial f(x' + \alpha (x - x'))}{\partial x_i} \, d\alpha.
\vspace{-1mm}
\end{equation}
IG accumulates gradients along a straight-line path from the baseline \( x' \) to the actual input \( x \), ensuring that attributions satisfy desirable properties like sensitivity and implementation invariance. This makes IG suitable for explaining high-dimensional models in image and text domains.

\vspace{-1mm}
\noindent$\rhd$~\textbf{Occlusion} (\citet{zeiler2014visualizing})
is a perturbation-based method that measures the change in the model's output when parts of the input are occluded. For a feature $i$, the attribution is computed as:
\vspace{-1mm}
\begin{equation}
\text{Occlusion}_i = f(x) - f(x^{-i}),
\vspace{-1mm}
\end{equation}
where $x^{-i}$ represents the input with feature $i$ occluded (e.g., replaced by a baseline value). Occlusion is particularly effective for visualizing feature importance in images and is also applicable to tabular data.

\begin{figure}[!t]
    \centering
    \includegraphics[width=0.9\linewidth]{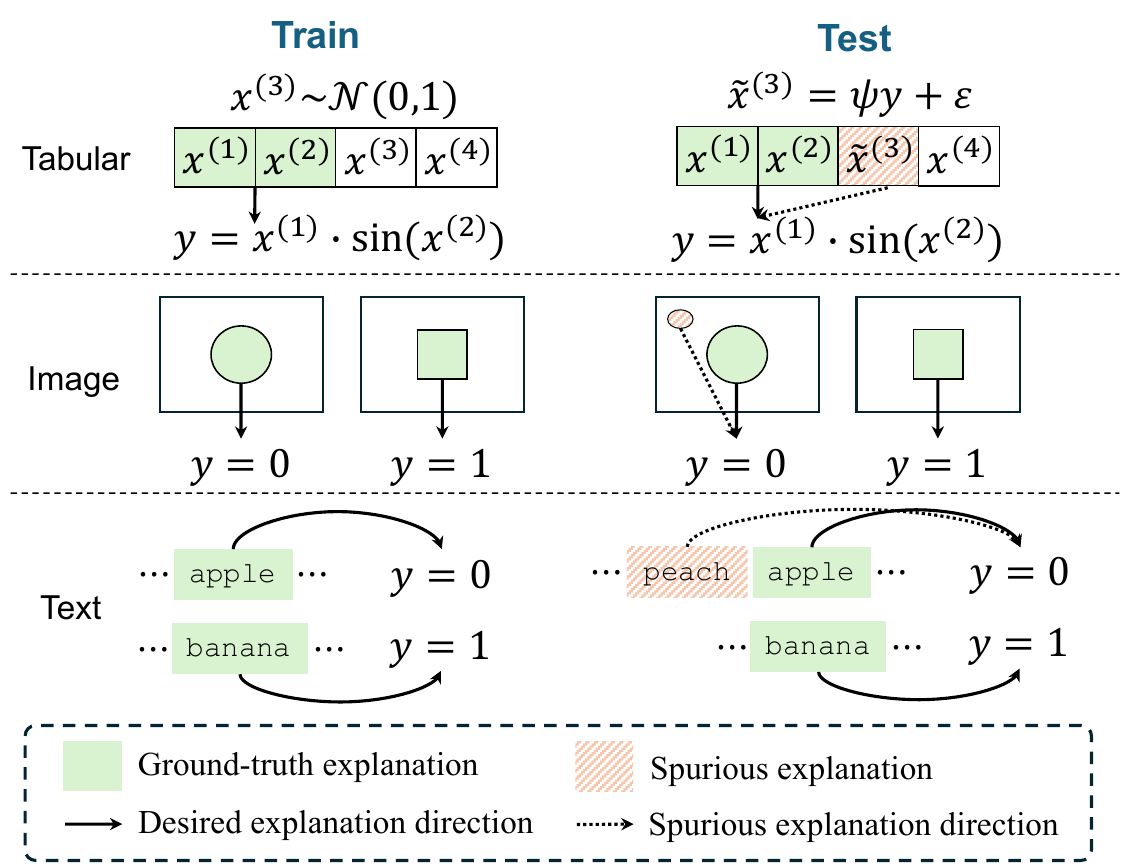}
    \vspace{-1mm}
    \caption{
    Illustration of IQ with spurious feature injection and its impact on explanations across modalities. 
For tabular data (first row), during training, feature $x^{(3)}$ follows a standard normal distribution, making it independent of the target variable. At test time, a spurious correlation is introduced where $x^{(3)}$ is linearly dependent on $y$ with noise $\varepsilon$, leading to incorrect reliance. 
For image data (second row), a distractor is injected into the test set, shifting explanations toward irrelevant regions. 
For text data (third row), an additional token (e.g., \texttt{peach}) appears in test samples, causing explanations to assign importance to non-informative words. 
    }
    \label{fig:spurious_explanation}
    \vspace{-4mm}
\end{figure}

\vspace{-1mm}
\section{Explanatory Inversion Quantification (IQ)}
\label{sec:iq}

In this section, we present the theoretical foundations of the proposed \emph{Inversion Quantification} framework for evaluating post-hoc explanations. We first formalize explanatory inversion and its implications for model interpretability (\S\ref{subsec:explanatory_inversion}). Then, we derive the components of IQ (\S\ref{subsec:iq_components}) and establish its applicability in both synthetic and real-world contexts.

\subsection{Explanatory Inversion}
\label{subsec:explanatory_inversion}

Let $\mathbf{x} \in \mathbb{R}^d$ denote the input vector, $M: \mathbb{R}^d \to \mathbb{R}^k$ represent a machine learning model mapping inputs to outputs, and $\mathcal{E}: \mathbb{R}^d \to \mathbb{R}^d$ denote a post-hoc explanation method assigning attributions $\mathbf{a} = \mathcal{E}(\mathbf{x})$. A reliable explanation should reflect the forward relationship between $\mathbf{x}$ and $M(\mathbf{x})$. However, explanatory inversion occurs when the attributions $\mathbf{a}$ primarily rely on the output $M(\mathbf{x})$, rather than capturing the input-output relationship.

\vspace{-1mm}
\begin{definition}\label{def:ei}
\textbf{[Explanatory Inversion]}
We define \emph{explanatory inversion} as the degree to which an explanation method $\mathcal{E}$ depends on the model's output $M(\mathbf{x})$ to generate attributions. The attributions $\mathbf{a}$ are expressed as:
\vspace{-1mm}
\begin{equation}
\mathbf{a} = f_{\mathcal{E}}(M(\mathbf{x}), \mathbf{x}),
\vspace{-1mm}
\end{equation}
where $f_{\mathcal{E}}$ is the internal mechanism of the explanation method. Explanatory inversion is maximal when $\mathbf{a}$ is independent of $\mathbf{x}$, i.e., $\|\partial \mathbf{a} / \partial \mathbf{x} \|\approx 0$, and minimal when $\|\partial \mathbf{a} / \partial M(\mathbf{x}) \|\approx 0$.
\end{definition}

\vspace{-3mm}
\subsection{Inversion Quantification}
\label{subsec:iq_components}
\vspace{-1mm}
Inversion Quantification (IQ) evaluates explanations along two key dimensions: reliance on outputs and explanation faithfulness. Intuitively, if the explanation does not rely on the outputs and aligns well with the model's decision-making process,  this explanation is considered less likely to involve explanatory inversion.
In the following, we first quantify the reliance on outputs ($R$) and the explanation faithfulness $F$, which are closely related to explanatory inversion. After that, we define the Inversion Score (IS) as a metric for IQ, combining $R$ and $F$.

\begin{definition}\label{def:r}\textbf{[Reliance on Outputs, $R$]}
The reliance score quantifies the degree to which the attributions $\mathbf{a}$ are influenced by the model's output $M(\mathbf{x})$, rather than the input-output relationship. For a dataset $\mathcal{D} = \{(\mathbf{x}_i, M(\mathbf{x}_i))\}_{i=1}^N$, $R$ is defined as:
\vspace{-1mm}
\begin{equation}
R = \frac{1}{N} \sum_{i=1}^N \frac{1}{d} \sum_{j=1}^d \rho\bigl(\Delta a_i^{(j)}, \Delta M(\mathbf{x}_i; j)\bigr),
\vspace{-1mm}
\end{equation}
where $d$ is the number of features, $\rho$ is the correlation coefficient, $\Delta a_i^{(j)} = a_i^{(j)} - a_{\text{base},i}^{(j)}$ represents the change in the attribution for feature $j$ after perturbation, where $a_{\text{base},i}^{(j)}$ is the baseline attribution for feature $j$, and $\Delta M(\mathbf{x}_i; j)$ is the observed change in the model’s output due to perturbing feature $j$ in input $\mathbf{x}_i$. A higher $R$ indicates stronger reliance on $M(\mathbf{x})$ and less explanatroy inversion, vice versa.
\end{definition}


\begin{definition}\label{def:f}{\textbf{[Explanation Faithfulness, $F$]}}
Faithfulness evaluates how well the attributions $\mathbf{a}$ align with the actual effect of features on the model's output. For a dataset $\mathcal{D} = \{(\mathbf{x}_i, M(\mathbf{x}_i))\}_{i=1}^N$, $F$ is defined as:
\begin{equation}
F = \frac{1}{N} \sum_{i=1}^N \frac{\sum_{j=1}^d a_i^{(j)} \big| M(\mathbf{x}_i) - M(\mathbf{x}_i^{(-j)}) \big|}{\sum_{j=1}^d |a_i^{(j)}|},
\end{equation}
where $a_i^{(j)}$ is the attribution assigned to feature $j$ for sample $\mathbf{x}_i$, and $M(\mathbf{x}_i^{(-j)})$ represents the model's output with feature $j$ perturbed or masked. 
\end{definition}

According to the definition, higher $F$ indicates that the attributions more faithfully reflect the effect of features on the model's output, thus relatively suffering less from explanatory inversion. Based on the definitions of $R$ and $F$, we define the inversion score (IS) as follows:
\begin{definition}\label{def:is}\textbf{[Inversion Score, $IS$]}
The Inversion Score (IS) quantifies the extent of explanatory inversion by combining Reliance on Outputs ($R$) and Explanation Faithfulness ($F$) into a single metric. IS is defined as:
\begin{equation}
\mathrm{IS}(R, F) = \left( \frac{R^p + (1-F)^p}{2} \right)^{\frac{1}{p}},
\end{equation}
where $p$ is a hyperparameter controlling the sensitivity to deviations in $R$ and $F$. By default, $p=2$ yields a quadratic mean $\mathrm{IS} \in [0, 1]$, which balances the contributions of $R$ and $F$. An ideal explanation with $R=0, F=1$ will lead to $\mathrm{IS} = 0$. An explanation purely based on inversion will have $R=1, F=0$ will lead to $\mathrm{IS} = 1$.

\textbf{Justification.} The power mean with $p>1$ emphasizes larger deviations in either $R$ or $(1-F)$, ensuring that significant shortcomings in one dimension dominate more for the score. This design reflects the intuition that both low faithfulness and high reliance on outputs severely undermine the quality of explanations, and their combined effect should be penalized. The choice of $p$ provides flexibility to adjust this sensitivity to the application.
\end{definition}

To justify the effectiveness of our proposed inversion scores measuring explanatory inversion, we prove the theorem:
\begin{theorem}
\label{thm:is_measures_inversion}
The proposed Inversion Score (IS) effectively quantifies explanatory inversion as defined in Definition \ref{def:ei}. The proof is presented in Appendix~\ref{app:is_measures_inversion_proof}. Specifically:
\vspace{-3mm}
\begin{enumerate}[leftmargin=*]
    \item Higher IS indicates stronger reliance on the model's output $M(\mathbf{x})$ (captured by $R$) and weaker alignment of attributions $\mathbf{a}$ with the causal effects of features on $M(\mathbf{x})$ (captured by $F$).
    \vspace{-2mm}
    \item Lower IS reflects explanations that minimize reliance on $M(\mathbf{x})$ and maximize faithfulness to the forward relationship between $\mathbf{x}$ and $M(\mathbf{x})$.
\end{enumerate}
\end{theorem}
\vspace{-2mm}
According to Theorem~\ref{thm:is_measures_inversion}, \methodname~can effectively measure the degree of explanatory inversion presented in explanations. In addition, in scenarios where ground-truth feature importance $\mathbf{g}$ is available (e.g., synthetic datasets), we define the alignment score $A$ as follows:
%
%
\begin{definition}\label{def:cosine}{\textbf{[Explanation Alignment, $A$]}} Given the attribution vector $\mathbf{a}$ and the ground-truth vector $\mathbf{g}$, the explanation alignment $A$ is calculated as their cosine similarity:
\begin{equation}
A = \frac{\langle \mathbf{a}, \mathbf{g} \rangle}{\|\mathbf{a}\| \|\mathbf{g}\|}.
\end{equation}
This definition applies across tabular, image, and text data by adapting $\mathbf{g}$ appropriately (e.g., causal features, bounding boxes, or ground-truth tokens).
\end{definition}


\begin{figure*}[!t]
    \centering
    \includegraphics[width=0.99\linewidth]{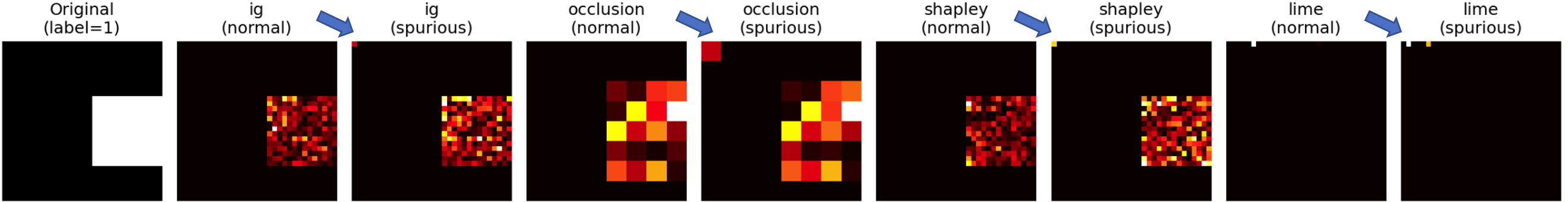}
    \caption{Visualization of feature attributions for the shape classification task under both normal and spurious scenarios. Each row displays an input image and feature attributions generated by four post-hoc explanation methods: Integrated Gradients (IG), Occlusion, Shapley Value Sampling, and LIME. Columns show comparisons between normal (left) and spurious (right) conditions. The spurious scenario introduces a bright distractor pixel in the top-left corner of images labeled as 1, which leads to incorrect attributions in several methods. Desired focus regions (e.g., the object shape) are highlighted under normal conditions, while spurious conditions shift the attributions toward the irrelevant injected pixel. 
    More case studies are included in Appendix~\ref{app:case}.
    }
    \label{fig:tease_exp}
    \vspace{-4mm}
\end{figure*}

\vspace{-3mm}
\subsection{Spurious Feature Injection}
\label{subsec:spurious}
As shown in figure~\ref{fig:spurious_explanation}, to evaluate the robustness of explanation methods against explanatory inversion, we introduce a spurious feature $\widetilde{x}_{\mathrm{spur}}$ that is correlated with the model's output. This feature is injected \textbf{only during inference time}, ensuring that the decision-making learned by the model during training remains unchanged. $\widetilde{x}_{\mathrm{spur}}$ is defined as:
\begin{equation}\label{eq:spur}
\widetilde{x}_{\mathrm{spur}} = \psi M(\mathbf{x}) + \varepsilon, \quad \varepsilon \sim \mathcal{N}(0, \sigma^2),
\end{equation}
where $\psi$ controls the strength of the correlation between $\widetilde{x}_{\mathrm{spur}}$ and the model's output $M(\mathbf{x})$, and $\varepsilon$ is random noise.

\textbf{Motivation.} By design, this injection ensures that the model's decision boundary and predictions remain unaffected, but it provides a new feature that is heavily correlated with the prediction, and explanation methods may erroneously attribute importance to this new feature. A good post-hoc explanation method should maintain the original explanation and not assign undue importance to the spurious feature. Any changes in the explanation reflect the degree of explanatory inversion.
The impact of the spurious feature on explanation methods is assessed by measuring the difference in the Inversion Score (IS) before and after introducing $\widetilde{x}_{\mathrm{spur}}$, denoted as $\Delta \mathrm{IS}$:
\vspace{-2mm}
\begin{equation}
\Delta \mathrm{IS} = \mathrm{IS}_{\mathrm{spur}} - \mathrm{IS}_{\mathrm{base}},
\vspace{-2mm}
\end{equation}
where $\mathrm{IS}_{\mathrm{base}}$ is the IS computed on the original dataset, and $\mathrm{IS}_{\mathrm{spur}}$ is the IS computed after injecting the spurious feature. A larger $\Delta \mathrm{IS}$ indicates that the explanation method is more susceptible to spurious correlations, reflecting its vulnerability to explanatory inversion.

\vspace{-2mm}
\section{Reproduce-by-Poking (\methodname)}
\label{sec:rbp}

To address the issue of explanatory inversion, we propose \emph{Reproduce-by-Poking (\methodname)}, a novel enhancement to post-hoc explanation methods. \methodname~incorporates forward perturbation checks into the attribution process, encouraging that attributions reflect genuine input-output relationships rather than artifacts of model outputs. This section describes the design of \methodname~and provides an intuitive and mathematical justification for its effectiveness.

\vspace{-2mm}
\subsection{Framework of \methodname}
\methodname~refines standard post-hoc explanations through forward perturbation checks, introducing an additional validation step.
First, we compute the baseline attributions, denoted as $\mathbf{a}$, using any established explanation method $\mathcal{E}$ (e.g., SHAP, LIME, IG, or Occlusion). These baseline attributions reflect the initial assessment of how each feature in the input $\mathbf{x}$ influences the model’s output. 

\textbf{Forward Perturbation Checks.}  
For each sample $\mathbf{x}$ and its feature $j$, \methodname~perturbs the feature multiple times to generate a series of modified inputs, $\mathbf{x}_{\text{pert},j}$. The goal is to simulate slight variations in the feature values while maintaining an unaltered prediction, i.e., $M(\mathbf{x}_{\text{pert},j}) = M(\mathbf{x})$. At each perturbation, new attributions, $\mathbf{a}_{\text{pert},j}$, are computed. The deviation for feature $j$ is then quantified as:
\begin{equation}
\delta^{(j)} = \frac{1}{n_{\text{pert}}} \sum_{k=1}^{n_{\text{pert}}} \big| a_{\text{pert}(k)}^{(j)} - a^{(j)} \big|,
\end{equation}
where $a_{\text{pert}(k)}^{(j)}$ represents the attribution for feature $j$ after the $k$-th perturbation, and $a^{(j)}$ is the baseline attribution. This step captures the stability of the attributions under small, controlled changes to the input. Features with high deviations are indicative of attributions that are overly sensitive to minor perturbations, suggesting potential unreliability.

\begin{figure}[t]
    \centering
\includegraphics[width=1\linewidth]{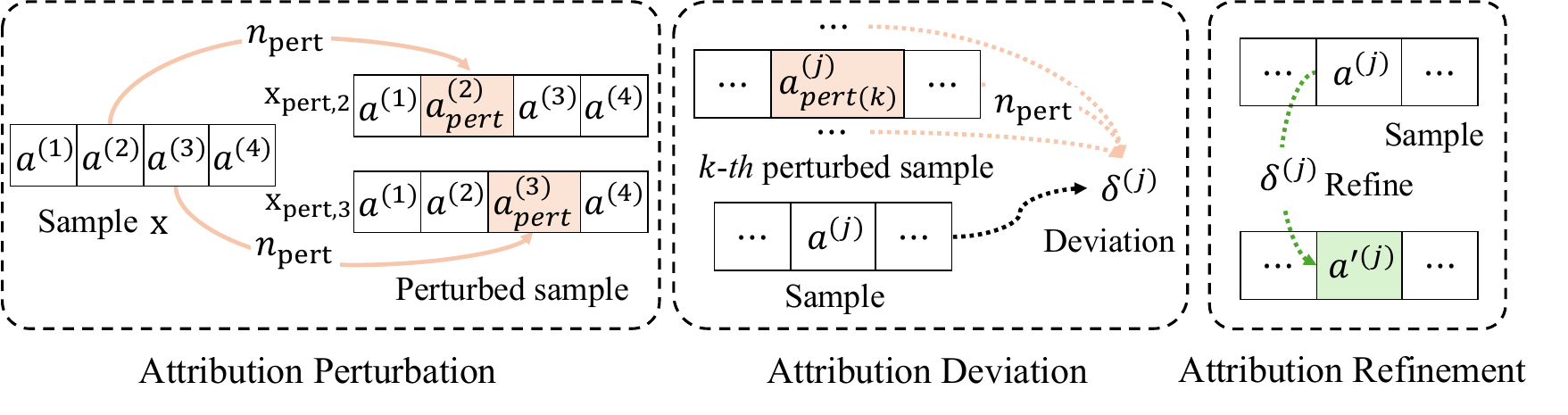}
        \caption{Overview of \methodname, divided into three stages. In the \emph{Attribution Perturbation} stage (left), multiple perturbed samples are generated from a given input sample $\mathbf{x}$, altering feature values. In the \emph{Attribution Deviation} stage (middle), deviations $\delta^{(j)}$ are computed for each feature $a^{(j)}$ based on differences across perturbations. Finally, in the \emph{Attribution Refinement} stage (right), attributions are refined by reducing the influence of features with high deviation, yielding adjusted attributions $a^{\prime {(j)}}$. 
        }
    \label{fig:RBP_framework}
    \vspace{-3mm}
\end{figure}

\vspace{-1mm}
\textbf{Adjust Baseline Attributions.}
In the final step, \methodname~refines the initial attributions by penalizing features with high deviations. The refined attributions, $\tilde{{a}}$, are computed as:
\begin{equation}
\tilde{{a}}^{(j)} = \frac{a^{(j)}}{1 + \delta^{(j)} \cdot \lambda},
\end{equation}
where $\lambda > 0$ is a hyperparameter that controls the impact of the deviation on the adjustment. This adjustment reduces the influence of features whose attributions are unstable, ensuring that the refined attributions are more robust.


\textbf{Intuition} behind forward perturbation in \methodname~is that reliable attributions should remain stable under small, localized changes to input features. If the attribution for a feature fluctuates significantly despite the model's prediction staying constant, it indicates that the attribution is overly sensitive and may not reflect the true causal importance of the feature. Conversely, stable attributions under such perturbations suggest that the feature's importance is accurately captured, aligning with the model's actual behavior. By penalizing unstable attributions, \methodname~ensures explanations are more robust and faithful to the model's underlying mechanisms.

\subsection{Theoretical Analysis}
\label{subsec:theoretical_properties}

We present two key theoretical properties of \methodname~that demonstrate its ability to reduce explanatory inversion, and a theorem to indicate the robustness of \methodname~against spurious features. Proofs of the theorems are in Appendix~\ref{app:proof}.

\begin{theorem}[Reduction of Output Reliance]\label{thm:reduction_reliance}
The reliance score $R_{RBP}$ achieved by \methodname~is strictly smaller than the reliance score $R$ of the baseline method, i.e.,
\vspace{-2mm}
\begin{equation}
R_{RBP} < R,
\vspace{-2mm}
\end{equation}
indicating a reduction in explanatory inversion by penalizing attributions that fail to align with forward perturbations.
\end{theorem}
\begin{theorem}[Faithfulness Improvement]\label{thm:faithfulness_improvement}
The faithfulness score $F_{RBP}$ achieved by \methodname~is strictly smaller than the faithfulness score $F$ of the baseline method, i.e.,
\vspace{-2mm}
\begin{equation}
F_{RBP} > F,
\vspace{-2mm}
\end{equation}
indicating improved alignment of attributions with true feature effects through perturbation-driven adjustments.
\end{theorem}
Theorem~\ref{thm:reduction_reliance} and Theorem~\ref{thm:faithfulness_improvement} demonstrate that \methodname~is able to reduce explanatory inversion by reducing reliance on output and improving explanation faithfulness. To demonstrate the performance of \methodname~in the presence of spurious features, we further prove the following theorem:
\begin{theorem}[Resilience to Spurious Features]\label{thm:resilience_spurious}
Let $\widetilde{x}_{\text{spur}}$ be a spurious feature introduced during inference according to Eq.\eqref{eq:spur}. If $\widetilde{x}_{\text{spur}}$ does not causally influence $M(\mathbf{x})$, \methodname~ensures that the adjusted attribution $\tilde{\mathbf{a}}_{\text{spur}}$ for the spurious feature converges to zero:
\vspace{-1mm}
\begin{equation}
\tilde{\mathbf{a}}_{\text{spur}} \to 0 \quad \text{if } \delta_{\text{spur}} \gg 0 \text{ or } \Delta M(\mathbf{x}; \widetilde{x}_{\text{spur}}) \approx 0.
\vspace{-2mm}
\end{equation}
\end{theorem}
\vspace{-2mm}
According to Theorem~\ref{thm:resilience_spurious}, \methodname~is able to produce adjusted attribution that converges to zero when the feature is spurious.
In summary, \methodname~mitigates explanatory inversion by grounding attributions in forward perturbations, reducing reliance on the model’s outputs and increasing the overall reliability of post-hoc explanations.

\vspace{-2mm}
\section{Experiments}
\label{sec:experiments}
\vspace{-1mm}

For comprehensiveness, our evaluation spans across multiple data \textbf{modalities}, \textit{i.e.}, tabular, image, and text, under controlled settings where artificially introduced \emph{spurious features} may influence post-hoc explanations. We consider both regression and classification \textbf{tasks}, and cover a range of \textbf{models}, including traditional machine learning methods (\textit{e.g.}, support vector machines and random forests) and deep learning architectures (\textit{e.g.}, multilayer perceptrons (MLPs), convolutional neural networks (CNNs), and transformers).

\vspace{-1mm}
\subsection{Experimental Setup}
\vspace{-1mm}

Each dataset consists of a training set and two test sets: a \textbf{standard} test set and a \textbf{spurious} test set, where an additional irrelevant feature is introduced at inference time. The evaluation metrics include \emph{reliance on outputs} ($R$), \emph{faithfulness} ($F$), the overall \emph{Inversion Score} ($IS$), the change in IS due to spurious features ($\Delta IS$), and \emph{alignment} ($A$). 
To ensure fair comparisons, we confirm that all models converge during training and achieve high predictive performance on the original test set (see Appendix~\ref{app:implementation}, \ref{app:train} for implementation details). This ensures that models correctly capture the intended data relationships, allowing us to isolate the effects of spurious feature injection on explanation methods. Experimental results on real-world datasets are in Appendix~\ref{app:cifar}. 

\vspace{-3mm}
\hrulefill\\
$\rhd$\,\textbf{Synthetic\,Tabular\,Data}:\,Multi-Feature\,Regression. We construct a synthetic regression dataset with six input features: $\{x^{(1)}, x^{(2)}\}$ are informative, while $\{x^{(3)}, x^{(4)}, x^{(5)},\\ x^{(6)}\}$ serve as distractors. The datasets are generated according to the following randomly chosen non-linear function. 
Similar patterns are observed with other functions.
\vspace{-1mm}
\begin{equation}
y = x^{(1)} \cdot \sin(x^{(1)}) \cdot \log(1 + |x^{(2)}|) + \varepsilon.
\vspace{-1mm}
\end{equation}
\textbf{Spurious Feature Injection.} 
In the spurious test set, we transform one of the dummy feature by $\tilde{x}^{(j)} = \psi y + \epsilon, (j \in \{3,4,5,6\})$, which is correlated with the target variable at test time but is not used for learning during training. This simulates a setting where models may inadvertently rely on spurious correlations in explanations.

\vspace{-1mm}
\textbf{Models.} 
We evaluate a range of regression models, including:
Random Forest Regressor, {Support Vector Regression} (with RBF kernel), and {MLP}.


\vspace{-3mm}
\hrulefill\\
$\rhd$\,\textbf{Synthetic\,Image\,Data}:\,Shape\,Classification.
We generate a dataset of $32\times32$ grayscale images, where each image contains either a \textbf{circle} (label = $0$) or a \textbf{square} (label = $1$). The shapes are randomly positioned within the image, and each sample is annotated with a bounding box marking the region corresponding to the shape.

\vspace{-1mm}
\textbf{Spurious Pixel Injection.} 
In the spurious test set, a bright pixel is introduced in the top-left corner of all images labeled $y=1$, creating a distractor that may mislead the explanation.

\vspace{-1mm}
\textbf{Model.}
We train a CNN with two convolutional layers followed by fully connected layers. We also provide results of ResNets in Appendix~\ref{app:cifar}.


\vspace{-3mm}
\hrulefill\\
$\rhd$\,\textbf{Synthetic\,Text\,Data}:\,Keyword-Based\,Classification.
We construct a binary classification dataset where a sentence is labeled $y=1$ if it contains the keywords ``\texttt{apple}'' or ``\texttt{banana}'', and $y=0$ otherwise. This setup allows for a straightforward evaluation of whether explanations correctly highlight the relevant tokens. Note that the setting is slightly different from the illustration in Figure~\ref{fig:intro-post-hoc},\ref{fig:spurious_explanation}, which are slightly modified for better illustration.

\textbf{Spurious Token Injection.} 
In the spurious test set, a non-informative token (e.g., ``\texttt{peach}'') is inserted into sentences with $y=1$ with 95\% probability. This tests whether the model incorrectly attributes importance to the inserted token instead of the true predictive ones.

\textbf{Model.} 
We fine-tune a {TinyBERT}-based classifier~\cite{jiao2020tinybert}, a lightweight transformer with two layers.






\begin{table*}[t]
    \caption{Quantitative evaluation of explanation methods under normal and spurious scenarios across tabular, image, and text datasets. The scores are averaged over 3 random seeds and reported as percentages (\%). For each method, we measure $R$ (Reliance), $F$ (Faithfulness), $\mathrm{IS}$ (Inversion Score), and $A$ (Alignment) for both baseline and \methodname. Additionally, results are provided for spurious test sets, where \textbf{$\Delta \mathrm{IS}$} represents the change in inversion score between baseline and \methodname. The results highlight that \methodname~consistently reduces the inversion score across all datasets and explanation methods, and indicate improved robustness against spurious features. See case studies in Appendix~\ref{app:case}.}\label{tab:main}
    \vspace{-2mm}
    \centering
    \small
    \setlength{\tabcolsep}{3pt} 
    \renewcommand{\arraystretch}{1.2} 
    \scalebox{0.79}{    
    \begin{tabular}{l l cccc cccc cccc cccc cc}
        \toprule
        \multirow{2}{*}{\textbf{Model}} & \multirow{2}{*}{\textbf{Explanation}} & \multicolumn{4}{c}{\textbf{Baseline}} & \multicolumn{4}{c}{\textbf{\methodname}} & \multicolumn{4}{c}{\textbf{Spurious Baseline}} & \multicolumn{4}{c}{\textbf{Spurious \methodname}} & \textbf{Baseline} & \textbf{\methodname} \\
        \cmidrule(lr){3-6} \cmidrule(lr){7-10} \cmidrule(lr){11-14} \cmidrule(lr){15-18}
        &  & $R$ $(\downarrow)$ & $F$ $(\uparrow)$ & $\mathrm{IS}$ $(\downarrow)$ & $A$ $(\uparrow)$& $R$ $(\downarrow)$ & $F$ $(\uparrow)$ & $\mathrm{IS}$ $(\downarrow)$ & $A$ $(\uparrow)$& $R$ $(\downarrow)$ & $F$ $(\uparrow)$ & $\mathrm{IS}$ $(\downarrow)$ & $A$ $(\uparrow)$& $R$ $(\downarrow)$ & $F$ $(\uparrow)$ & $\mathrm{IS}$ $(\downarrow)$ & $A$ $(\uparrow)$ & $\Delta$IS $(\downarrow)$  &  $\Delta$IS $(\downarrow)$ \\
        \midrule
        \rowcolor{yellow!20}\multicolumn{20}{c}{\textbf{Tabular (Nonlinear Multi-Feature Regression)}} \\
        \midrule
        Random  & SHAP & 8.7 & 67.3 & 24.0 & 88.6 & 7.9 & 81.7 & 14.1 & 92.2 & 11.4 & 47.3 & 38.2 & 83.3 & 9.7 & 60.5 & 28.7 & 87.8 & 14.2 & 14.6 \\
        Forest & LIME & 8.4 & 70.3 & 21.8 & 85.9 & 6.6 & 87.3 & 10.1 & 87.5 & 12.8 & 45.9 & 39.9 & 81.4 & 11.1 & 59.4 & 29.8 & 84.9 & 17.5 & 19.7 \\
        \midrule
        \multirow{2}{*}{SVM-RBF} & SHAP & 7.5 & 67.2 & 23.8 & 88.2 & 7.1 & 78.3 & 16.2 & 90.0 & 11.8 & 46.4 & 38.8 & 84.7 & 10.2 & 60.0 & 29.2 & 86.3 & 15.1 & 13.0 \\
        & LIME & 8.8 & 67.2 & 24.0 & 87.9 & 7.1 & 77.5 & 16.7 & 91.1 & 10.9 & 43.7 & 40.5 & 82.6 & 11.3 & 55.1 & 32.7 & 89.2 & 16.5 & 16.1  \\
        \midrule
        \multirow{4}{*}{MLP} & SHAP & 8.7 & 64.8 & 25.7 & 88.4 & 6.5 & 74.7 & 18.5 & 92.8 & 12.6 & 41.5 & 42.3 & 83.3 & 11.1 & 55.3 & 32.6 & 90.1 & 16.6 & 14.1 \\
        & LIME & 8.0 & 62.9 & 26.8 & 89.1 & 8.1 & 76.2 & 17.8 & 92.5 & 11.7 & 39.5 & 43.6 & 85.2 & 9.8 & 53.3 & 33.8 & 90.2 & 16.7 & 16.0 \\
         & IG & 9.2 & 62.0 & 27.6 & 90.3 & 6.9 & 73.7 & 19.2 & 92.4 & 12.5 & 38.1 & 44.7 & 84.9 & 11.0 & 54.1 & 33.4 & 87.8 & 17.0 & 14.2 \\
         & Occlusion & 11.1 & 64.2 & 26.5 & 91.3 & 6.6 & 74.3 & 18.8 & 93.7 & 12.1 & 40.7 & 42.8 & 86.5 & 11.0 & 53.9 & 33.5 & 90.9 & 16.3 & 14.7 \\
        \midrule
        \rowcolor{green!10}\multicolumn{20}{c}{\textbf{Image (Shape Classification)}} \\
        \midrule
        \multirow{4}{*}{CNN} & SHAP & 17.4 & 36.8 & 46.4 & -- & 9.6 & 82.0 & 14.4 & -- & 32.0 & 17.1 & 62.8 & -- & 14.2 & 65.0 & 26.7 & -- & 16.5 & 12.3 \\
         & LIME & 19.5 & 26.5 & 53.8 & -- & 15.8 & 35.6 & 46.9 & -- & 35.5 & 19.7 & 62.1 & -- & 17.4 & 21.6 & 56.8 & -- & 8.3 & 9.9 \\
         & IG & 11.3 & 38.8 & 44.0 & -- & 7.8 & 43.8 & 40.1 & -- & 19.3 & 25.6 & 54.4 & -- & 11.6 & 29.8 & 50.3 & -- & 10.3 & 10.2 \\
         & Occlusion & 21.6 & 39.3 & 45.6 & -- & 14.0 & 42.5 & 41.9 & -- & 29.5 & 22.8 & 58.4 & -- & 20.4 & 34.2 & 48.7 & -- & 12.9 & 6.9 \\
         \midrule
        \rowcolor{blue!10}\multicolumn{20}{c}{\textbf{Text (Keyword-based Classification)}} \\
        \midrule
        \multirow{3}{*}{TinyBERT} & SHAP & 7.4 & 14.2 & 60.9 & -- & 1.1 & 16.0 & 59.4 & -- & 46.00 & 11.0 & 70.8 & -- & 32.0 & 19.0 & 61.6 & -- & 10.0 & 2.2 \\
         & LIME & 3.2 & 18.0 & 58.0 & -- & 2.2 & 24.3 & 53.6 & -- & 5.4 & 17.5 & 58.5 & -- & 6.4 & 21.6 & 55.6 & -- & 0.4 & 2.1 \\
         & IG & 8.9 & 12.4 & 62.3 & -- & 4.3 & 16.8 & 58.9 & -- & 20.1 & 5.9 & 68.0 & -- & 14.7 & 11.2 & 63.7 & -- & 5.8 & 4.7 \\
        \bottomrule
    \end{tabular}
    }
    \label{tab:performance}
    \vspace{-4mm}
\end{table*}

\subsection{Main Results}
\vspace{-1mm}

Table~\ref{tab:main} presents the results for quantified explanatory inversion based on the proposed IQ framework. Based on the results, we draw the following key insights.

\vspace{-1mm}
\textbf{Explanatory inversion exists in all tasks, modalities, models, and explanation methods.}  
    The baseline inversion scores ($\mathrm{IS}$) are non-zero across all experiments, indicating that all explanation methods exhibit varying degrees of explanatory inversion, regardless of data modality (tabular, image, or text), model architecture (e.g., random forest, CNN, TinyBERT), or explanation method (e.g., SHAP, LIME, IG). The results also exhibit that explanatory inversion varies across models and tasks, which means it is hard to judge which post-hoc explanation method will suffer less from the inversion for a given model or task. However, we find that if the task is extremely simple, the inversion diminishes, as shown in Table~\ref{tab:linear_tab_performance} in Appendix~\ref{app:linear_tab}. 

\vspace{-1mm}
\textbf{Explanatory inversion is amplified by the injection of a spurious feature correlated with the output prediction.}  
    The spurious feature injection during inference leads to a consistent increase in $R$, a decrease in $F$, and an overall increase in the inversion score ($\mathrm{IS}$). This pattern is observed across all models and explanation methods. For example, in tabular tasks, $\mathrm{IS}$ increases by more than $10\%$ under spurious conditions, validating that explanatory inversion intensifies.

\vspace{-1mm}
\textbf{Increased inversion is associated with decreased alignment between attributions and ground-truth feature importance.}  
    A clear negative correlation between $\mathrm{IS}$ and $A$ is observed. As $\mathrm{IS}$ increases under spurious conditions, $A$ consistently declines across models and explanation methods. This indicates that explanatory inversion not only reduces the reliability of the explanation but also worsens the alignment of attributions with true feature importance. 

\vspace{-1mm}
\textbf{\methodname~consistently mitigates explanatory inversion.}  
    The application of \methodname~reduces the reliance score ($R$), increases faithfulness ($F$), and lowers the inversion score ($\mathrm{IS}$). Moreover, \methodname~results in a smaller $\Delta \mathrm{IS}$, indicating that explanation methods become more robust to spurious correlations. This is evidenced by the reduced $\Delta \mathrm{IS}$ values for \methodname~compared to the baseline across all modalities. Specifically, the average reduction in $\Delta \mathrm{IS}$ is approximately $0.95\%$ for tabular data, $2.19\%$ for image data, and $2.39\%$ for text data. This leads to a $1.8\%$ improvement across all the explanation methods and domains. These findings confirm that \methodname~enhances robustness against spurious correlations, particularly for tasks involving textual and image data.

\begin{figure*}[t]
    \centering
\includegraphics[width=1\linewidth]{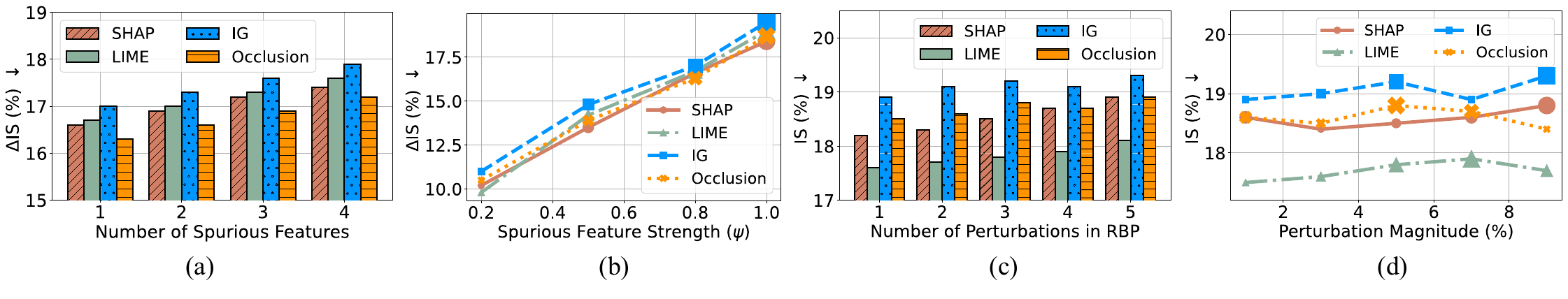}
\vspace{-8mm}
    \caption{Ablation study analyzing (\textbf{i}). the properties of explanatory inversion via different spurious injection settings of IQ; and (\textbf{ii}) robustness and effectiveness of \methodname~under various hyper-parameter conditions. \textbf{(a)} Impact of varying the number of spurious features on $\Delta \mathrm{IS}$. \textbf{(b)} Influence of spurious feature strength $\psi$ on $\Delta \mathrm{IS}$. \textbf{(c)} Effect of the number of perturbation check in \methodname~on IS. \textbf{(d)} Influence of perturbation magnitude in \methodname~on $\mathrm{IS}$.}
    \label{fig:ablation4}
    \vspace{-4mm}
\end{figure*}

\vspace{-1mm}
\subsection{Impact from Various Spurious Feature Injection}
\label{subsec:ablation_spur}
\vspace{-1mm}

To further analyze the properties of \emph{explanatory inversion}, particularly focusing on spurious feature injection, we conduct a series of experiments to evaluate two key parameters in the IQ framework: (a) the number of spurious features injected and (b) the strength of the spurious feature ($\psi$).
The experiment is conducted on the synthesized tabular data with MLP as the task model. 

As shown in Figure~\ref{fig:ablation4} (a), increasing the number of spurious features leads to a consistent rise in $\Delta \mathrm{IS}$ across all explanation methods. This behavior aligns with the theoretical understanding that more spurious features can confuse the attribution process by creating multiple competing false explanations. Besides, we observe similar trends among different numbers of injected spurious features. By default, we set the number to 1 for simplicity and better illustration.

\vspace{-1mm}
In Figure~\ref{fig:ablation4} (b), for a given spurious feature, we vary the spurious feature strength ($\psi$), which controls the degree of correlation between the injected spurious feature and the model's output. As $\psi$ increases, $\Delta \mathrm{IS}$ rises for all methods, indicating stronger explanatory inversion due to more pronounced spurious correlations. By default, we
set the $\psi=0.8$ for major experiments.
\vspace{-2mm}
\subsection{Parameter Sensitivity Analyses of \methodname}
\label{subsec:para-rbp}
\vspace{-1mm}

To evaluate the sensitivity of \methodname~to different parameter choices, based on Section~\ref{sec:rbp}, on the tabular dataset with MLP model, we conduct experiments on two key parameters: (c) the number of perturbations applied in \methodname~and (d) the magnitude of perturbation noise. These parameters control the robustness and stability of \methodname. 

As shown in Figure~\ref{fig:ablation4} (c), increasing the number of perturbations in \methodname~has a relatively stable impact on $\mathrm{IS}$ across all explanation methods. While slight fluctuations are observed, particularly for the IG and SHAP methods, the overall stability suggests that \methodname~is robust to variations in the number of perturbations. Importantly, methods like Occlusion exhibit minor changes, indicating that the number of perturbations does not significantly degrade performance under these settings. In practice, we perform 3 perturbations for \methodname.

In Figure~\ref{fig:ablation4} (d), we vary the perturbation noise magnitude from $1\%$ to $9\%$. The results show that \methodname~maintains consistent IS values across this range, with only minor variations. This indicates that \methodname~is not overly sensitive to the exact magnitude of noise used for forward perturbations. The stability of $\mathrm{IS}$ across different noise levels highlights the adaptability of \methodname~in diverse scenarios. The results confirm that \methodname~balances perturbation-based refinement without introducing additional significant instability in attribution results. We use $5\%$ magnitude for \methodname~by default.

\begin{figure}[t]
    \centering
    \includegraphics[width=1\linewidth]{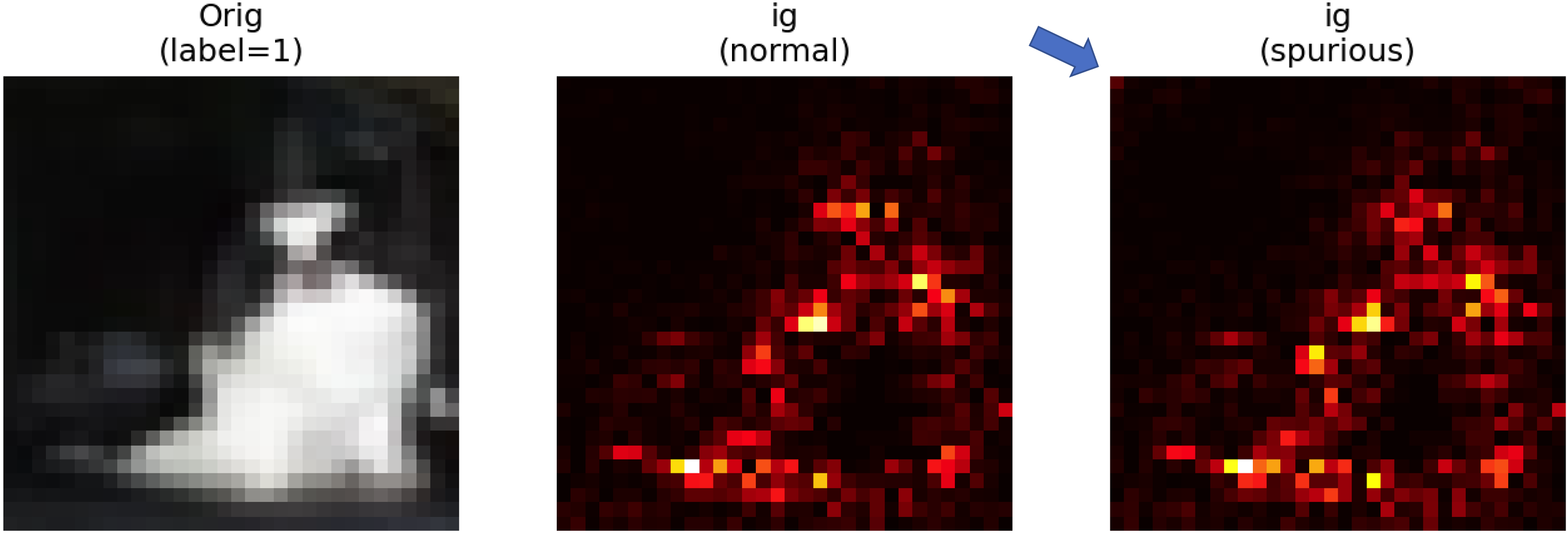}
    \vspace{-5mm}
    \caption{A case study of using IG for explaining a prediction from ResNet-18~\cite{he2016deep} on the CIFAR Dataset. We observe that the injected pixel distractor is highlighted (you might need to enlarge the figure to see). 
    See more results in Appendix~\ref{app:cifar}.}
    \label{fig:cifar_main}
    \vspace{-2mm}
\end{figure}

\vspace{-1mm}
\subsection{Generalization to Real-World Applications}

The proposed IQ framework for quantifying explanatory inversion is validated through a case study on a real-world dataset. Figure~\ref{fig:cifar_main} illustrates the behavior of IG when applied to a ResNet-18 model on a CIFAR image correctly classified as a ``dog''. Under normal conditions, the attribution map highlights relevant areas of the image, including key features of the dog. However, when a spurious distractor pixel is injected, the explanation shifts toward this irrelevant feature, demonstrating an increased susceptibility to explanatory inversion. 
This case study reveals that explanatory inversion is not limited to synthetic datasets or controlled scenarios but also manifests in real-world applications.

\vspace{-1mm}
\section{Conclusion}
\label{sec:conclusion}
\vspace{-1mm}
In this work, we introduce the \emph{Inversion Quantification} (\emph{IQ}) framework to measure \emph{explanatory inversion}, where post-hoc explanations rely on model outputs rather than input-output relationships. Our analyses reveal that widely used methods like LIME and SHAP suffer from this issue across tabular, image, and text domains, especially under spurious correlations.
To address this, we propose \emph{Reproduce-by-Poking} (\methodname), which integrates forward perturbation checks to reduce inversion. We show the effectiveness of \methodname~theoretically and empirically. Our findings emphasize the importance of reliable explanation methods in real-world applications. Future research will explore expanding \methodname~to handle multi-modal models and complex settings while maintaining scalability and computational efficiency. 

\newpage
\section*{Impact Statements}
This paper presents work whose goal is to advance the field of Machine Learning. There are many potential societal consequences of our work, none of which we feel must be specifically highlighted here.

\bibliography{example_paper}
\bibliographystyle{icml2025}

\newpage
\appendix
\onecolumn
\section{Discussion on Newer Methods for Post-hoc Explanations}\label{app:related}

Post-hoc explanation methods for machine learning models can be broadly categorized into several groups based on their underlying techniques and application domains.
(1) Gradient-Based Explanation Methods: These methods leverage gradient information from deep learning models to highlight influential features contributing to a prediction. Examples include Grad-CAM \cite{selvaraju2017grad}, Score-CAM \cite{wang2020score}, and Attention Flow in Transformers \cite{abnar2020quantifying}.
(2) Perturbation-Based Explanation Methods: These methods modify the input and analyze changes in the model’s output to infer feature importance. Various works, including Rise \cite{petsiuk2018rise}, S-LIME \cite{zhou2021s}, DLIME \cite{zafar2019dlime}, and DSEG-LIME \cite{knab2024dseg}, are proposed to improve stability, determinism, and segmentation in image-based explanations.
(3) Counterfactual Explanations: These methods provide interpretability by generating alternative scenarios or feature importance scores. Examples include DiCE~\cite{mothilal2020explaining}, CAL~\cite{rao2021counterfactual}, TalkToModel~\cite{slack2023explaining}, and DVCEs~\cite{augustin2022diffusion}.
(4) Feature Attribution Methods: These methods propose to improve classic post-hoc feature attribution methods, including TreeSHAP~\cite{lundberg2017consistent}, Kernel SHAP~\cite{lundberg2017unified}, Extended Kernel SHAP~\cite{aas2021explaining}, along with the extension of SHAP to various scenarios~\cite{fryer2021shapley,van2022tractability, kumar2020problems}.

In terms of explanations on natural language, recent advancements have proposed to explain language models in a post-hoc manner~\cite{ding2022explainability,kaur2022trustworthy, kroeger2023large, krishna2024post}. Regarding the trustworthiness of post-hoc explanations, researchers have proposed to assess the effectiveness and fairness of explanation methods~\cite{slack2021reliable, dai2022fairness, adebayo2022post, slack2020fooling}. More recently, researchers have considered incorporating human feedback to improve the quality of post-hoc explanations~\cite{bianchi2024interpretable, jesus2021can, han2022explanation, agarwal2021towards}.

As a prosperous field, exhaustively experimenting with all post-hoc methods is impractical. In this paper, we choose the 4 most widely used methods, on which most of the later methods are built. Also, we only test on tabular, image, and text data. Comprehensive study on other modalities, including multi-modal senarios, are potential direction for future works.

\section{Contrast to Previous Evaluation of Post-hoc Explanations}

Evaluating the reliability and faithfulness of post-hoc explanations generated by methods like LIME, SHAP, and Integrated Gradients (IG) is a crucial area of XAI research. A significant body of work focuses on assessing whether explanations accurately reflect the model's internal reasoning process -- often termed ``faithfulness''.

\subsection{Established Faithfulness Metrics}

Several metrics have been proposed to quantify faithfulness.

\textit{Completeness}, as demonstrated by Integrated Gradients, ensures that attributions sum up to the difference between the model's output for the input and a baseline.

\textit{Sensitivity} metrics assess how attributions change in response to input perturbations. For instance, sensitivity in IG relates to the axiom that if an input differs from the baseline in only one feature, and the model's prediction changes, that feature should receive non-zero attribution. Similarly, \citet{yeh2019fidelity} define sensitivity concerning the explanation's response to small input perturbations.

\textit{Perturbation-based} or \textit{Deletion/Insertion tests} measure the change in model output when features deemed important (or unimportant) by the explanation are removed or added. Methods like Occlusion inherently operate on this principle, and benchmarks like those by \citet{hooker2019benchmark} formalize this evaluation.

\textit{Infidelity}, proposed by \citet{yeh2019fidelity}, measures the expected squared difference between the dot product of the input perturbation and the explanation, and the change in the model's output due to that perturbation. It quantifies how well the explanation aligns with the model's local sensitivity to input changes.

\subsection{Robustness and Stability}

Another line of evaluation focuses on the robustness or stability of explanations, examining how much attributions change under small, often imperceptible, perturbations to the input~\cite{alvarez2018robustness}. Work by \citet{tan2023robust} explores the potential trade-offs between explanation robustness and faithfulness. While related to sensitivity, robustness specifically emphasizes the consistency and reliability of the explanation itself against minor input variations.

\subsection{Explanatory Inversion - A Complementary Perspective}

This paper introduces the concept of \textit{Explanatory Inversion}, which describes a failure mode where explanations become overly conditioned on the model's output, potentially rationalizing predictions post-hoc rather than reflecting the forward reasoning process from inputs to outputs. We propose the \textit{Inversion Quantification (IQ)} framework and the \textit{Inversion Score} $IS$ to measure this phenomenon. $IS$ combines two dimensions: \textit{Reliance on Outputs} $R$, which quantifies the correlation between attributions and model predictions under perturbation, and \textit{Explanation Faithfulness} $F$, which assesses alignment with the actual effect of features on the model's output using input perturbations.

Our proposed $IS$ metric offers a distinct perspective compared to existing faithfulness evaluations:

\begin{itemize}
    \item \textbf{$IS$ vs.\ Infidelity/Sensitivity:} While infidelity and sensitivity focus on the alignment between explanation attributions and the effects of input perturbations on model output, $IS$ (particularly through the $R$ component) directly assesses the correlation between attributions and the model's output itself. Explanatory Inversion tackles the directionality of the explanation process -- questioning if explanations are derived from the output rather than explaining how the output was reached. Therefore, $IS$ is conceptually different and not merely a special case of infidelity; it measures a distinct failure mode related to reversed justification.
    
    \item \textbf{$IS$ vs.\ Robustness/Stability:} Robustness measures the stability of attributions under input perturbations, whereas $IS$ measures the dependence of attributions on the output. While unstable explanations might exhibit higher inversion, the core focus differs; $IS$ specifically targets the potential for explanations to rationalize predictions.
    
    \item \textbf{$IS$ vs.\ Completeness/Deletion Tests:} Metrics like completeness and deletion tests verify if attributions account for the model's output difference or predict performance drops upon feature removal, respectively. They assess the forward impact assumption. $IS$, conversely, scrutinizes the potential reverse dependency -- whether attributions are primarily dictated by the output, a scenario particularly relevant when spurious correlations exist.
\end{itemize}

In summary, the proposed Inversion Score ($IS$) provides a novel and complementary evaluation axis within the XAI toolkit. It specifically quantifies the risk of ``explanatory inversion,'' where attributions reflect output correlations rather than the model's input-driven reasoning, complementing existing metrics that focus on input sensitivity alignment, attribution completeness, or stability.


    

\section{Proofs.}\label{app:proof}

\subsection{Proof of Theorem \ref{thm:is_measures_inversion}}
\label{app:is_measures_inversion_proof}

\textbf{Restating Inversion Quantification.}
Recall that explanatory inversion is defined as the degree to which attributions \(\mathbf{a}\) depend on the model's output \(M(\mathbf{x})\), rather than capturing the forward relationship between \(\mathbf{x}\) and \(M(\mathbf{x})\). To quantify this, the Inversion Score (IS) combines two components:
\begin{itemize}[leftmargin=*]
    \item \textbf{Reliance on Outputs (\(R\))}: Higher \(R\) indicates stronger reliance of attributions on \(M(\mathbf{x})\), reflecting potential backward explanations.
    \item \textbf{Explanation Faithfulness (\(F\))}: Higher \(F\) reflects better alignment between attributions and the causal effect of features on \(M(\mathbf{x})\).
\end{itemize}

The IS is defined as:
\begin{equation}
\mathrm{IS}(R, F) = \left( \frac{R^p + (1-F)^p}{2} \right)^{\frac{1}{p}},
\end{equation}
where \(p > 1\) is a hyperparameter controlling the sensitivity to deviations in \(R\) and \(F\).

\textbf{Proof of Dependency on \(\|\partial \mathbf{a} / \partial \mathbf{x}\|\).}
We aim to show that \(\mathrm{IS}(R, F) \propto \|\partial \mathbf{a} / \partial \mathbf{x}\|\). 

\begin{proof}
The reliance score \(R\) and faithfulness score \(F\) are defined as follows:
\begin{equation}
R = \frac{1}{N} \sum_{i=1}^N \frac{1}{d} \sum_{j=1}^d \rho(\Delta a_i^{(j)}, \Delta M(\mathbf{x}_i; j)),
\end{equation}
\begin{equation}
F = \frac{1}{N} \sum_{i=1}^N \frac{\sum_{j=1}^d a_i^{(j)} |\Delta M(\mathbf{x}_i; j)|}{\sum_{j=1}^d |a_i^{(j)}|}.
\end{equation}

\noindent Step-by-step, we analyze how \(R\) and \(F\) depend on \(\|\partial \mathbf{a} / \partial \mathbf{x}\|\):

\textbf{Step 1: Behavior of \(R\).}
The reliance score \(R\) measures the correlation \(\rho(\Delta a_i^{(j)}, \Delta M(\mathbf{x}_i; j))\), where \(\Delta a_i^{(j)} = a_i^{(j)} - a_{\text{base}}^{(j)}\). The term \(\Delta a_i^{(j)}\) is directly influenced by changes in \(\mathbf{a}\) due to perturbations in \(\mathbf{x}\). Specifically:
\begin{equation}
\Delta a_i^{(j)} \propto \frac{\partial \mathbf{a}}{\partial \mathbf{x}},
\end{equation}
since small perturbations in \(\mathbf{x}\) cause proportional changes in the attributions \(\mathbf{a}\). Thus:
\begin{equation}
R \propto \|\partial \mathbf{a} / \partial \mathbf{x}\|.
\end{equation}

\textbf{Step 2: Behavior of \(F\).}
The faithfulness score \(F\) evaluates how well attributions align with the causal effect of features on \(M(\mathbf{x})\). By definition:
\begin{equation}
F = \frac{1}{N} \sum_{i=1}^N \frac{\sum_{j=1}^d a_i^{(j)} |\Delta M(\mathbf{x}_i; j)|}{\sum_{j=1}^d |a_i^{(j)}|}.
\end{equation}
Here:
\begin{equation}
\Delta M(\mathbf{x}_i; j) = M(\mathbf{x}_i) - M(\mathbf{x}_i^{(-j)}),
\end{equation}
and \(\Delta M(\mathbf{x}_i; j)\) is influenced by how sensitive the model's output is to changes in feature \(j\). The attributions \(a_i^{(j)}\), however, depend on \(\mathbf{x}\) through:
\begin{equation}
a_i^{(j)} \propto \frac{\partial \mathbf{a}}{\partial \mathbf{x}}.
\end{equation}
Hence, \(F\) indirectly depends on \(\|\partial \mathbf{a} / \partial \mathbf{x}\|\), as the quality of attributions is influenced by how changes in \(\mathbf{x}\) align with the observed effects \(\Delta M(\mathbf{x}_i; j)\).

\textbf{Step 3: Dependency of \(\mathrm{IS}(R, F)\) on \(\|\partial \mathbf{a} / \partial \mathbf{x}\|\).}
Combining the dependencies of \(R\) and \(F\) on \(\|\partial \mathbf{a} / \partial \mathbf{x}\|\), the Inversion Score \(\mathrm{IS}(R, F)\) becomes:
\begin{equation}
\mathrm{IS}(R, F) = \left( \frac{R^p + (1 - F)^p}{2} \right)^{\frac{1}{p}}.
\end{equation}
Since both \(R\) and \(1 - F\) are proportional to \(\|\partial \mathbf{a} / \partial \mathbf{x}\|\), we conclude that:
\begin{equation}
\mathrm{IS}(R, F) \propto \|\partial \mathbf{a} / \partial \mathbf{x}\|.
\end{equation}

\textbf{Conclusion.}
The Inversion Score \(\mathrm{IS}(R, F)\) directly reflects the sensitivity of attributions \(\mathbf{a}\) to changes in the input \(\mathbf{x}\). A higher \(\|\partial \mathbf{a} / \partial \mathbf{x}\|\) implies greater explanatory inversion, as captured by the combined metrics \(R\) and \(F\) in the \(\mathrm{IS}\) formulation. This thus complete the proof.
\end{proof}

\subsection{Proofs of Theoretical Properties}
\label{appendix:proofs}

\textbf{Theorem 5.1.}
\textit{
\methodname~reduces reliance on the model’s output, i.e.,
\begin{equation}
R_{RBP} < R.
\end{equation}
}

\begin{proof}
The reliance score for the baseline method is defined as:
\begin{equation}
R = 1 - \frac{1}{N} \sum_{i=1}^N \frac{1}{d} \sum_{j=1}^d \rho\bigl(\Delta a_i^{(j)}, \Delta M(\mathbf{x}_i; j)\bigr),
\end{equation}
where \(\Delta a_i^{(j)} = a_i^{(j)} - a_{\text{base}}^{(j)}\) captures deviations from the base attribution, and \(\Delta M(\mathbf{x}_i; j)\) is the observed change in the model’s output when feature \(j\) is perturbed.

For \methodname, the refined reliance score is:
\begin{equation}
R_{RBP} = 1 - \frac{1}{N} \sum_{i=1}^N \frac{1}{d} \sum_{j=1}^d \rho\bigl(\Delta \tilde{a}_i^{(j)}, \Delta M(\mathbf{x}_i; j)\bigr),
\end{equation}
where \(\Delta \tilde{a}_i^{(j)}\) are the refined attributions after incorporating perturbation-driven deviations.

The refined attributions are computed as:
\begin{equation}
\tilde{a}_i^{(j)} = \frac{a_i^{(j)}}{1 + \delta^{(j)} \cdot \lambda},
\end{equation}
where \(\delta^{(j)}\) captures the deviation observed under perturbations. Since features with high sensitivity to perturbations (\(\delta^{(j)}\)) are penalized, the refined attributions \(\Delta \tilde{a}_i^{(j)}\) become less influenced by noise or artifacts. This results in:
\begin{equation}
\rho\bigl(\Delta \tilde{a}_i^{(j)}, \Delta M(\mathbf{x}_i; j)\bigr) < \rho\bigl(\Delta a_i^{(j)}, \Delta M(\mathbf{x}_i; j)\bigr).
\end{equation}
Since correlation decreases, we obtain:
\begin{equation}
R_{RBP} < R.
\end{equation}
This completes the proof.
\end{proof}

\textbf{Theorem 5.2.}
\textit{
\methodname~improves faithfulness, i.e.,
\begin{equation}
F_{RBP} > F.
\end{equation}
}

\begin{proof}
The faithfulness score for the baseline method is defined as:
\begin{equation}
F = \frac{1}{N} \sum_{i=1}^N \frac{\sum_{j=1}^d a_i^{(j)} \big| \Delta M(\mathbf{x}_i; j) \big|}{\sum_{j=1}^d |a_i^{(j)}|}.
\end{equation}

For \methodname, the refined faithfulness score is:
\begin{equation}
F_{RBP} = \frac{1}{N} \sum_{i=1}^N \frac{\sum_{j=1}^d \tilde{a}_i^{(j)} \big| \Delta M(\mathbf{x}_i; j) \big|}{\sum_{j=1}^d |\tilde{a}_i^{(j)}|}.
\end{equation}

Since \methodname~adjusts attributions to reduce the influence of spurious or noisy attributions, the refined attributions satisfy:
\begin{equation}
\tilde{a}_i^{(j)} = \frac{a_i^{(j)}}{1 + \delta^{(j)} \cdot \lambda}.
\end{equation}

This adjustment aligns the attributions with the actual effect of features on the output, reducing faithfulness error. Since faithfulness measures alignment with true feature importance, the improvement in alignment due to \methodname~ensures that:
\begin{equation}
F_{RBP} > F.
\end{equation}
This completes the proof.
\end{proof}

\textbf{Theorem 5.3.}
\textit{
\methodname~ensures resilience to spurious features, i.e., for a spurious feature \(\widetilde{x}_{\text{spur}}\):
\begin{equation}
\tilde{a}_{\text{spur}} \to 0 \quad \text{if } \delta_{\text{spur}} \gg 0 \text{ or } \Delta M(\mathbf{x}; \widetilde{x}_{\text{spur}}) \approx 0.
\end{equation}
}

\begin{proof}
For a spurious feature \(\widetilde{x}_{\text{spur}}\), its baseline attribution \(a_{\text{spur}}\) is computed using a standard explanation method \(\mathcal{E}\):
\begin{equation}
a_{\text{spur}} = \mathcal{E}(\mathbf{x}, \widetilde{x}_{\text{spur}}).
\end{equation}

\methodname~refines this attribution by incorporating forward perturbation checks. Specifically, the deviation \(\delta_{\text{spur}}\) for \(\widetilde{x}_{\text{spur}}\) is calculated as:
\begin{equation}
\delta_{\text{spur}} = \frac{1}{n_{\text{pert}}} \sum_{k=1}^{n_{\text{pert}}} \big| a_{\text{pert},\text{spur}}^{(k)} - a_{\text{spur}} \big|,
\end{equation}
where \(a_{\text{pert},\text{spur}}^{(k)}\) is the attribution for \(\widetilde{x}_{\text{spur}}\) after the \(k\)-th perturbation.

If \(\widetilde{x}_{\text{spur}}\) is spurious, its attribution is expected to exhibit high sensitivity to perturbations due to its non-causal relationship with \(M(\mathbf{x})\). This results in:
\begin{equation}
\delta_{\text{spur}} \gg 0.
\end{equation}

Additionally, the causal effect of \(\widetilde{x}_{\text{spur}}\) on the model’s output, quantified as \(\Delta M(\mathbf{x}; \widetilde{x}_{\text{spur}})\), is negligible:
\begin{equation}
\Delta M(\mathbf{x}; \widetilde{x}_{\text{spur}}) \approx 0.
\end{equation}

The refined attribution \(\tilde{a}_{\text{spur}}\) is then computed as:
\begin{equation}
\tilde{a}_{\text{spur}} = \frac{a_{\text{spur}}}{1 + \delta_{\text{spur}} \cdot \lambda}.
\end{equation}

Since \(\delta_{\text{spur}} \gg 0\) or \(\Delta M(\mathbf{x}; \widetilde{x}_{\text{spur}}) \approx 0\), the refined attribution satisfies:
\begin{equation}
\tilde{a}_{\text{spur}} \to 0.
\end{equation}

Thus, \methodname~effectively eliminates attributions for spurious features, ensuring robustness to spurious correlations. This completes the proof.
\end{proof}

\section{Implementation Details}
\label{app:implementation}

In this section, we describe the implementation details for the experiments conducted across tabular, image, and text data domains. We outline dataset generation, model configurations, explanation methods, and key experimental setups.

\subsection{Tabular Data Experiments}
\textbf{Dataset:} We used a synthetic regression dataset with six features. One spurious feature $\tilde{x}_3$ was injected, defined as $\tilde{x}_3 = \psi M(\mathbf{x}) + \varepsilon$, where $\psi$ controls the spurious correlation strength, and $\varepsilon \sim \mathcal{N}(0, \sigma^2)$.  

\textbf{Models:} We evaluated three models:  
\begin{itemize}[leftmargin=*,itemsep=1pt]
    \item Random Forest (RF) with 100 trees.
    \item Support Vector Machine (SVM) with RBF kernel ($\gamma = 0.1$, $C = 1.0$).
    \item Multi-Layer Perceptron (MLP) with 2 hidden layers of size [128, 64].
\end{itemize}

\textbf{Explanation Methods:} Four post-hoc methods were applied: SHAP, LIME, Integrated Gradients (IG), and Occlusion. The explanations were evaluated under both normal and spurious conditions using the following metrics:  
\begin{itemize}[leftmargin=*,itemsep=1pt]
    \item Reliance on Outputs ($R$), Explanation Faithfulness ($F$), Inversion Score ($IS$), and Alignment ($A$).
    \item $\Delta IS$, measuring susceptibility to spurious correlations.
\end{itemize}

\textbf{Framework Implementation:}  
The \emph{Reproduce-by-Poking (RBP)} framework applied perturbations to each feature and computed deviations to refine attributions. Ablation studies were performed by varying the number of spurious features, the strength $\psi$, and the number of perturbations.

\subsection{Image Data Experiments}
\textbf{Dataset:} Shape classification was conducted on a synthetic dataset where images contained geometric shapes (e.g., circles, squares) with injected spurious distractor pixels in the top-left corner of images labeled as `1'.

\textbf{Models:} We used a CNN with 2 convolutional layers followed by 2 fully connected layers for the shape classification task.

\textbf{Explanation Methods:}  
We applied Integrated Gradients, Occlusion, Shapley Value Sampling, and LIME to generate visual attributions. These methods were evaluated using:  
\begin{itemize}[leftmargin=*,itemsep=1pt]
    \item $R$, $F$, $IS$, and alignment scores with ground-truth bounding boxes.
    \item Case studies to assess explanatory inversion on real-world image data.
\end{itemize}

\subsection{Text Data Experiments}
\textbf{Dataset:} We designed a keyword-based classification task where keywords were embedded in sentences. During inference, spurious distractor tokens highly correlated with the output labels were injected.

\textbf{Models:} A pre-trained TinyBERT model was fine-tuned for the classification task.

\textbf{Explanation Methods:} We applied SHAP, LIME, and Integrated Gradients to generate token-level attributions, which were evaluated based on:  
\begin{itemize}[leftmargin=*,itemsep=1pt]
    \item $R$, $F$, $IS$, and alignment with ground-truth keywords.
    \item The impact of varying numbers of distractor tokens and perturbation magnitudes.
\end{itemize}

We did not use Occlusion for text tasks, since it's usually used for tabular or image data.

\subsection{Reproduce-by-Poking (RBP) Parameters}
\textbf{Perturbation Process:} Features were perturbed three times per sample, with noise drawn from $\mathcal{N}(0, \sigma^2)$. Perturbation noise magnitudes ranged from $1\%$ to $9\%$ of feature values in sensitivity experiments.

\textbf{Hyperparameters:} 
\begin{itemize}[leftmargin=*,itemsep=1pt]
    \item The scaling factor $\lambda$ was set to 0.1 to refine attributions based on perturbation deviations.
    \item The power parameter $p$ in the Inversion Score was set to 2.
\end{itemize}

\textbf{Computational Resources:} Experiments were run on a machine with an NVIDIA RTX 3090 GPU and 64 GB of RAM. The code was implemented in Python using PyTorch, scikit-learn, and Captum for explanation generation.

\section{Training Dynamics.}\label{app:train}

\begin{figure*}[htbp]
    \centering
    \includegraphics[width=1\linewidth]{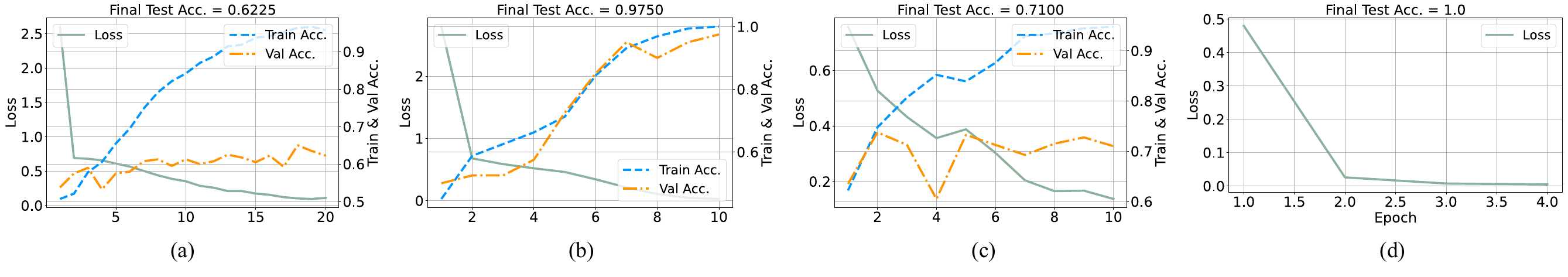}
    \caption{Train dynamics of (a) 2-layer CNN on CIFAR, (b) 2-layer CNN on synthesized shape classification dataset, (c) ResNet-18 on CIFAR, and (d) TinyBERT on synthesized keyword-based classification dataset.}
    \label{fig:train_dynamics}
\end{figure*}

When training the deep models for evaluating explanation methods, we use the common practice for training, validation, and test steps. The training dynamics are shown in Figure~\ref{fig:train_dynamics}. As observed, the task models have reasonable performance for evaluations.

\newpage
\section{Explanatory Inversion in Simple Linear Relationships}\label{app:linear_tab}

\begin{table*}[htpb]
    \caption{Results on Tabular data synthesized with a simple linear relationship. Scores are averaged with runs on 3 random seeds and reported as percentage ($\%$).}
    \centering
    \small
    \setlength{\tabcolsep}{3pt} 
    \renewcommand{\arraystretch}{1.2} 
    \scalebox{0.79}{    
    \begin{tabular}{l l ccc ccc}
        \toprule
        \multirow{2}{*}{\textbf{Model}} & \multirow{2}{*}{\textbf{Explanation}} & \multicolumn{3}{c}{\textbf{Baseline}}  & \multicolumn{3}{c}{\textbf{Spurious Baseline}} \\
        \cmidrule(lr){3-5} \cmidrule(lr){6-8} 
        &  & $R$ $(\downarrow)$ & $F$ $(\uparrow)$ & $\mathrm{IS}$ $(\downarrow)$ & $R$ $(\downarrow)$ & $F$ $(\uparrow)$ & $\mathrm{IS}$ $(\downarrow)$  \\
        \midrule
        \rowcolor{yellow!20}\multicolumn{8}{c}{\textbf{Tabular (Linear relationship: $y = x^{(1)} + x^{(2)} + \epsilon$)}} \\
        \midrule
        Random  & SHAP & 0 & 100.0 & 0  & 0 & 100.0 & 0  \\
        Forest & LIME & 0 & 100.0 & 0 & 0 & 100.0 & 0 \\
        \midrule
        \multirow{2}{*}{Linear Regression}  & SHAP & 0 & 100.0 & 0 &  0 & 100.0 & 0  \\
         & LIME & 0 & 100.0 & 0 & 0 & 100.0 & 0 \\
        \midrule
        \multirow{2}{*}{SVM}  & SHAP & 0 & 100.0 & 0 & 0 & 100.0 & 0  \\
         & LIME & 0 & 100.0 & 0 & 0 & 100.0 & 0 \\
        \midrule
        \multirow{4}{*}{MLP}  & SHAP & 0 & 98.5 & 1.1 & 0 & 98.0 & 1.4  \\
         & LIME & 0 & 99.8 & 0.2 & 0 & 97.3 & 1.9 \\
         & IG & 0 & 98.4 & 1.1 & 0 & 97.5 & 1.8 \\
         & Occlusion & 0 & 99.1 & 0.6 & 0 & 98.0 & 1.4 \\
        \bottomrule
    \end{tabular}
    }
    \label{tab:linear_tab_performance}
\end{table*}

Table~\ref{tab:linear_tab_performance} presents the results of explanatory inversion under a simple linear relationship between features and output, defined by \( y = x^{(1)} + x^{(2)} + \varepsilon \). In this setup, we observe the following key trends:

1. \textbf{Absence of Inversion in Most Models:}  
   For models such as Random Forest, Linear Regression, and SVM, both the reliance on outputs (\(R\)) and the Inversion Score (\(IS\)) are zero, indicating that these models produce faithful explanations without signs of explanatory inversion. The post-hoc methods (SHAP, LIME) provide explanations that align perfectly with the linear feature-output relationship, regardless of the presence of spurious features.

2. \textbf{Minimal Inversion for Neural Models:}  
   In contrast, MLP-torch exhibits minor signs of explanatory inversion. While \(R\) remains zero, the faithfulness score (\(F\)) decreases slightly under both baseline and spurious conditions. This leads to a small increase in \(IS\) (up to 1.9 for LIME), showing that even in simple linear settings, neural models may introduce subtle non-linear artifacts that affect post-hoc explanations.

3. \textbf{Spurious Features Have Negligible Effect:}  
   Across all methods and models, the spurious baseline results are virtually identical to the normal baseline. This highlights that explanatory inversion is strongly mitigated when the underlying feature-output relationship is linear, as post-hoc methods can accurately reflect the true feature contributions without being misled by spurious correlations.

Overall, these results suggest that explanatory inversion becomes less problematic when the model's decision function is simple and linear. This aligns with the theoretical expectation that explanatory inversion is exacerbated by complex, non-linear relationships and spurious influences.

\section{Case Studies}\label{app:case}

\begin{figure}[htpb]
    \centering
    \includegraphics[width=0.99\linewidth]{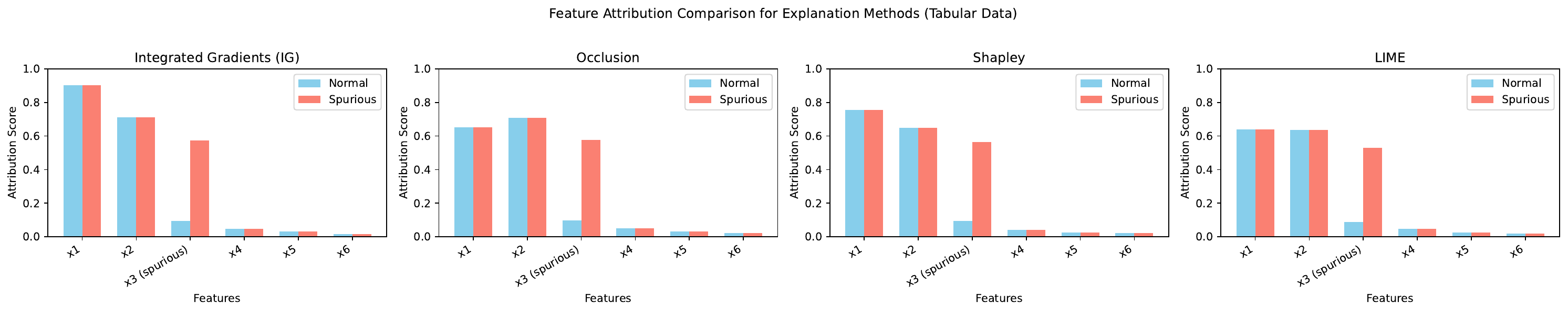}
    \caption{Feature attribution comparison for the multi-feature regression on tabular data across four explanation methods (Integrated Gradients, Occlusion, Shapley, and LIME) for tabular data. The bar charts display attributions for features $x_1$ to $x_6$ under both normal and spurious conditions. The spurious feature $x_3$ exhibits higher attribution under the spurious scenario, indicating a shift in explanation focus across all methods. While $x_1$ and $x_2$ maintain high relevance in normal conditions, the presence of a spurious correlation affects feature prioritization. This pattern demonstrates the potential susceptibility of explanation methods to spurious features.}
    \label{fig:tab}

\end{figure}

\begin{figure}[htpb]
    \centering
    \includegraphics[width=0.99\linewidth]{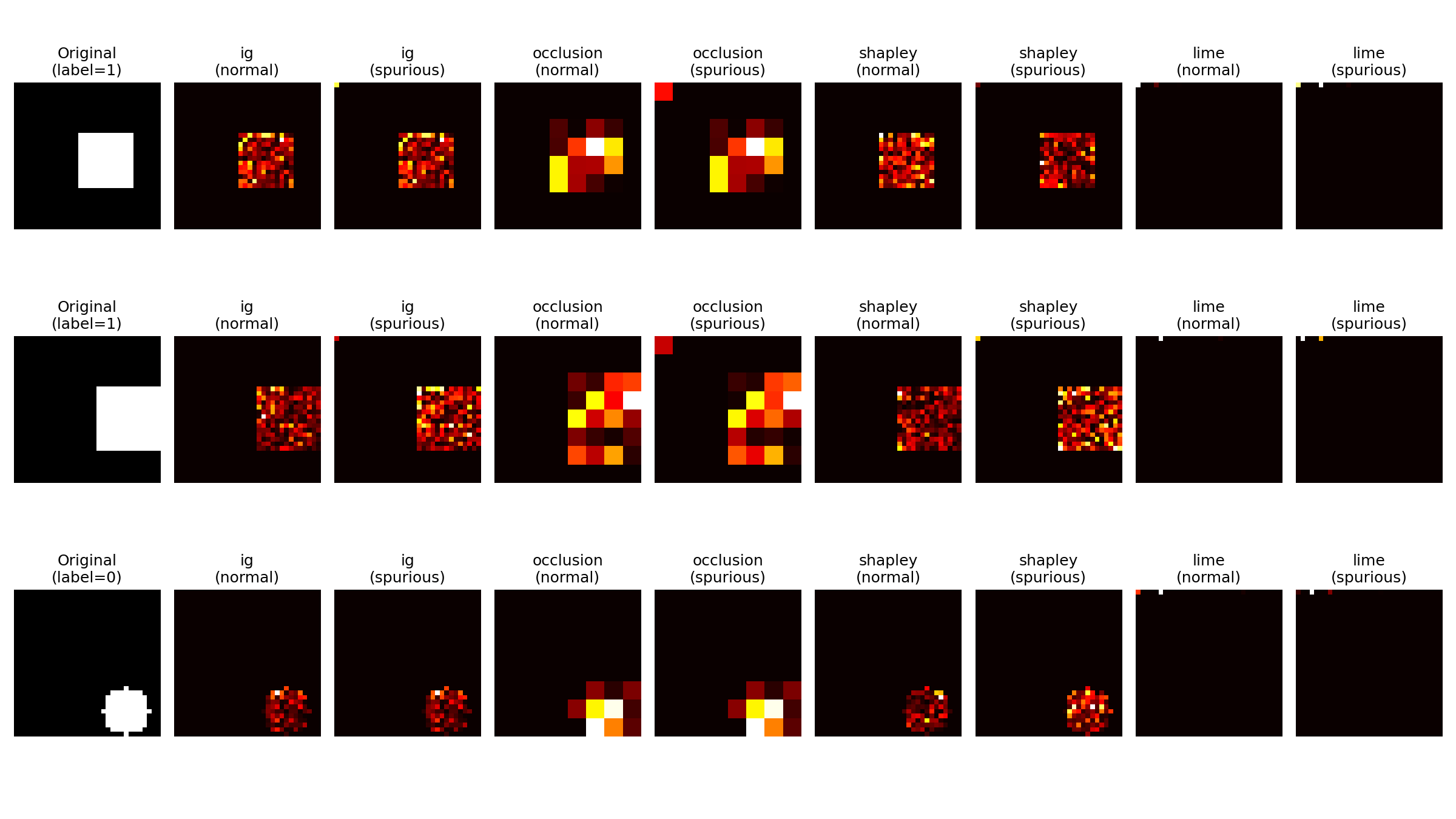}
    \caption{Visualization of feature attributions for the image classification task under both normal and spurious scenarios. Each row shows a different input image along with explanations generated by four post-hoc methods: Integrated Gradients (IG), Occlusion, Shapley Value Sampling, and LIME. Columns compare the attributions under normal conditions (middle) and spurious conditions (right). The spurious condition introduces a bright distractor pixel in the top-left corner, which shifts attributions toward the irrelevant region in several cases. The desired focus is highlighted by strong activations in relevant areas (e.g., shapes), while spurious influence results in increased attribution near the injected pixel.
    }
    \label{fig:img}
\end{figure}

\begin{figure}[htpb]
    \centering
    \includegraphics[width=0.99\linewidth]{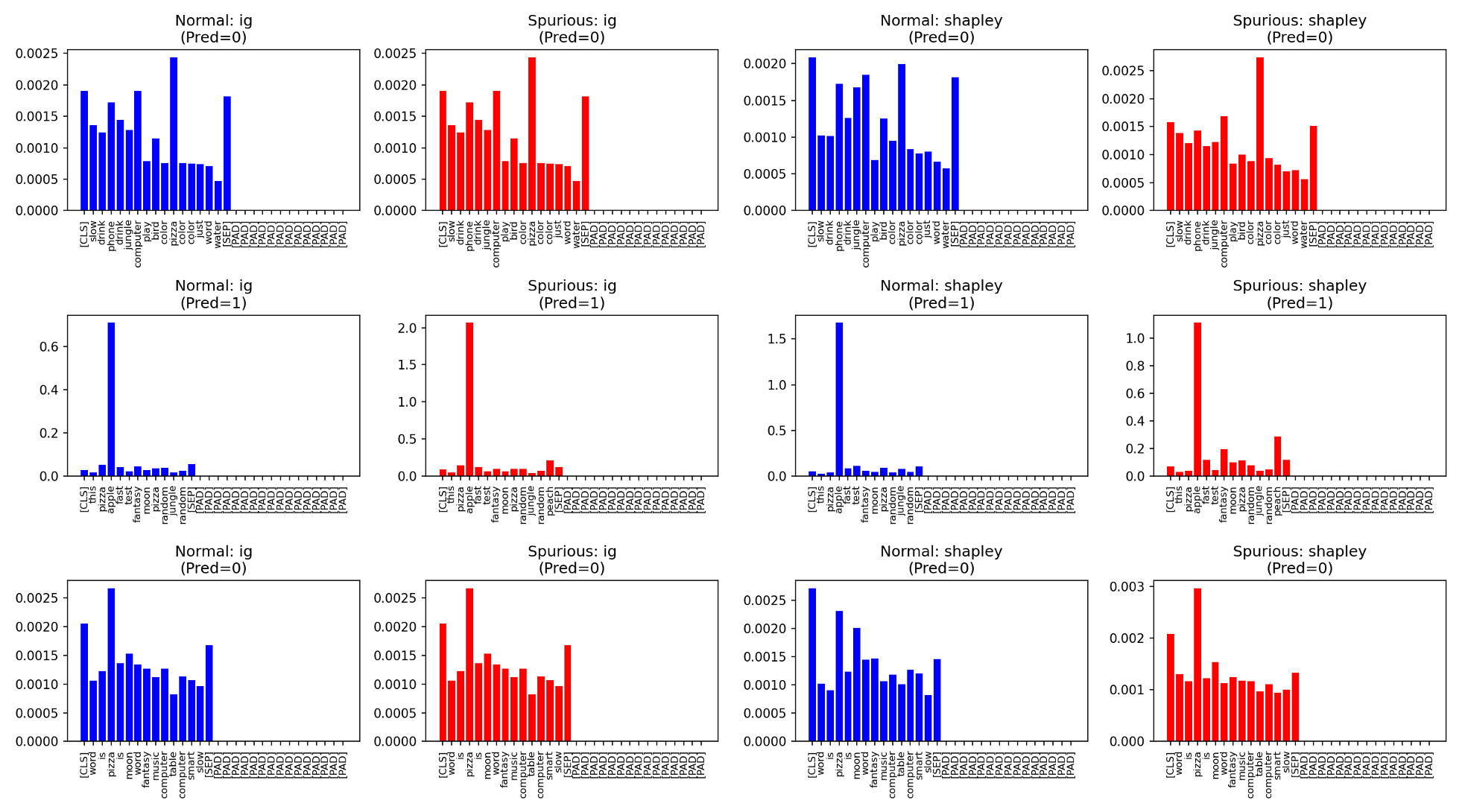}
        \caption{Feature attribution comparison on the text classification task under both normal and spurious conditions using two explanation methods: Integrated Gradients (IG) and Shapley Value Sampling. The horizontal axis shows tokenized input words, while the vertical axis represents attribution scores. The top rows illustrate results when the predicted label is $0$, and the bottom rows show predictions of $1$. Under spurious conditions (right panels), a spurious token (e.g., \texttt{peach}) affects the attribution scores, leading to a higher emphasis on non-relevant tokens. In contrast, under normal conditions (left panels), relevant tokens such as \texttt{apple} and \texttt{banana} maintain high attribution. This demonstrates the impact of spurious correlations on explanation consistency.}
    \label{fig:text}
\end{figure}

\section{Results on CIFAR dataset with ResNet-18}\label{app:cifar}

We conduct experiment on the CIFAR dataset. We select 2 classes (``cat'' $=$ $0$, ``dog'' $=$ $1$). We randomly sample 500 instances from the training set and 200 instances from the validation and test set. The model is trained with a classic image classification pipeline. Explanatory inversions can be observed on certain samples.

\begin{figure}[htpb]
    \centering
    \includegraphics[width=0.6\linewidth]{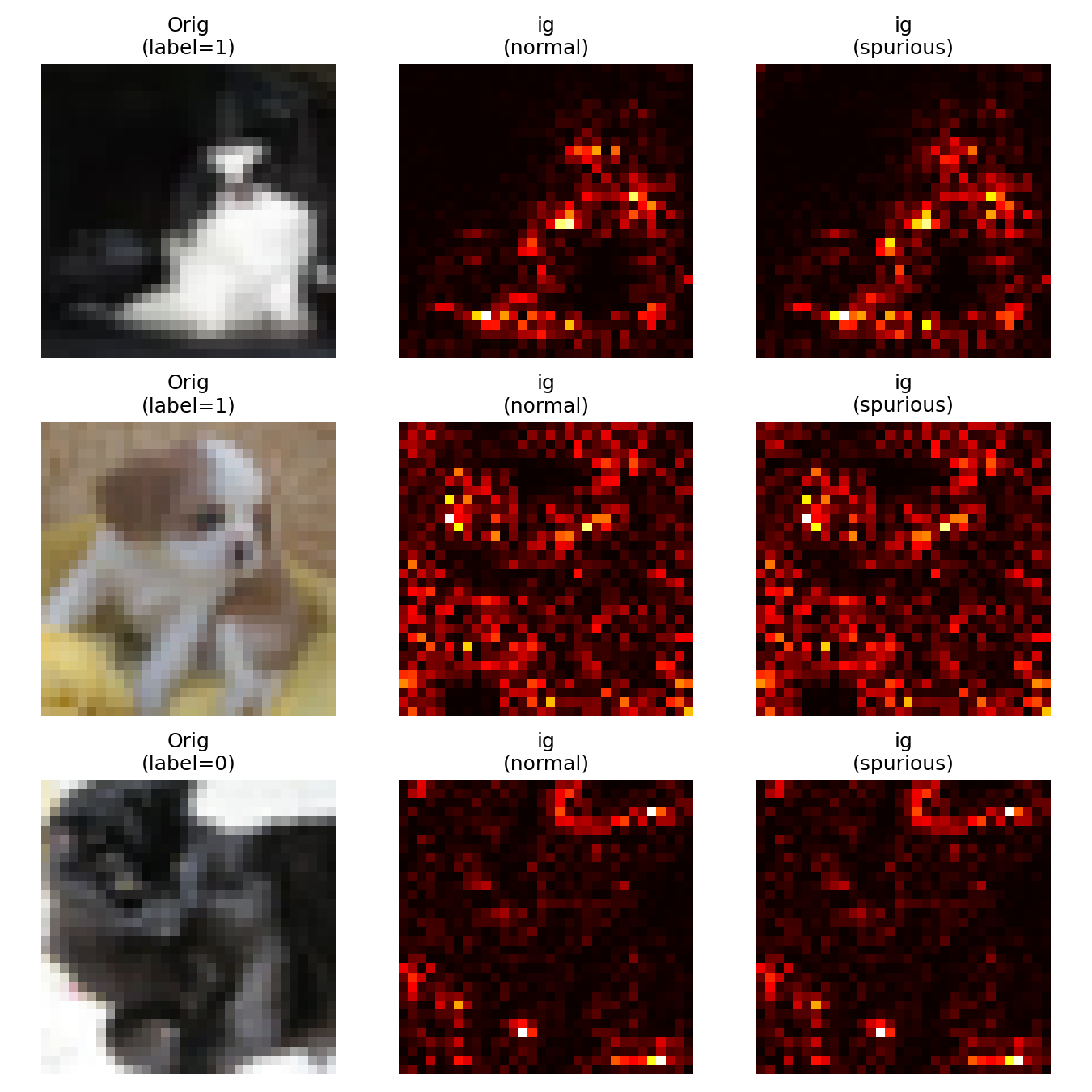}
        \caption{Case Studies on CIFAR}
\end{figure}

\begin{figure}[htpb]
    \centering
    \includegraphics[width=0.6\linewidth]{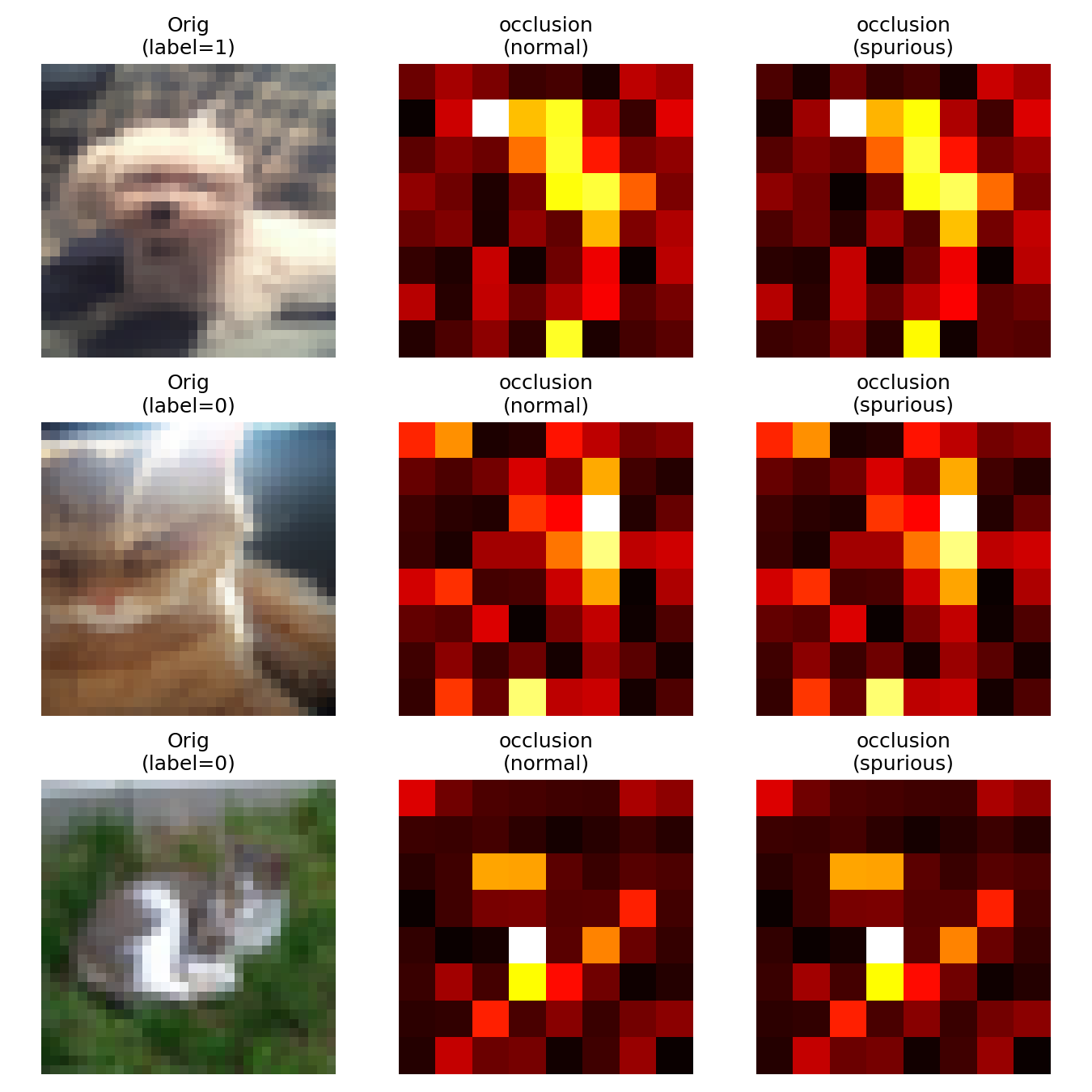}
        \caption{Case Studies on CIFAR}
\end{figure}

\begin{figure}[htpb]
    \centering
    \includegraphics[width=0.6\linewidth]{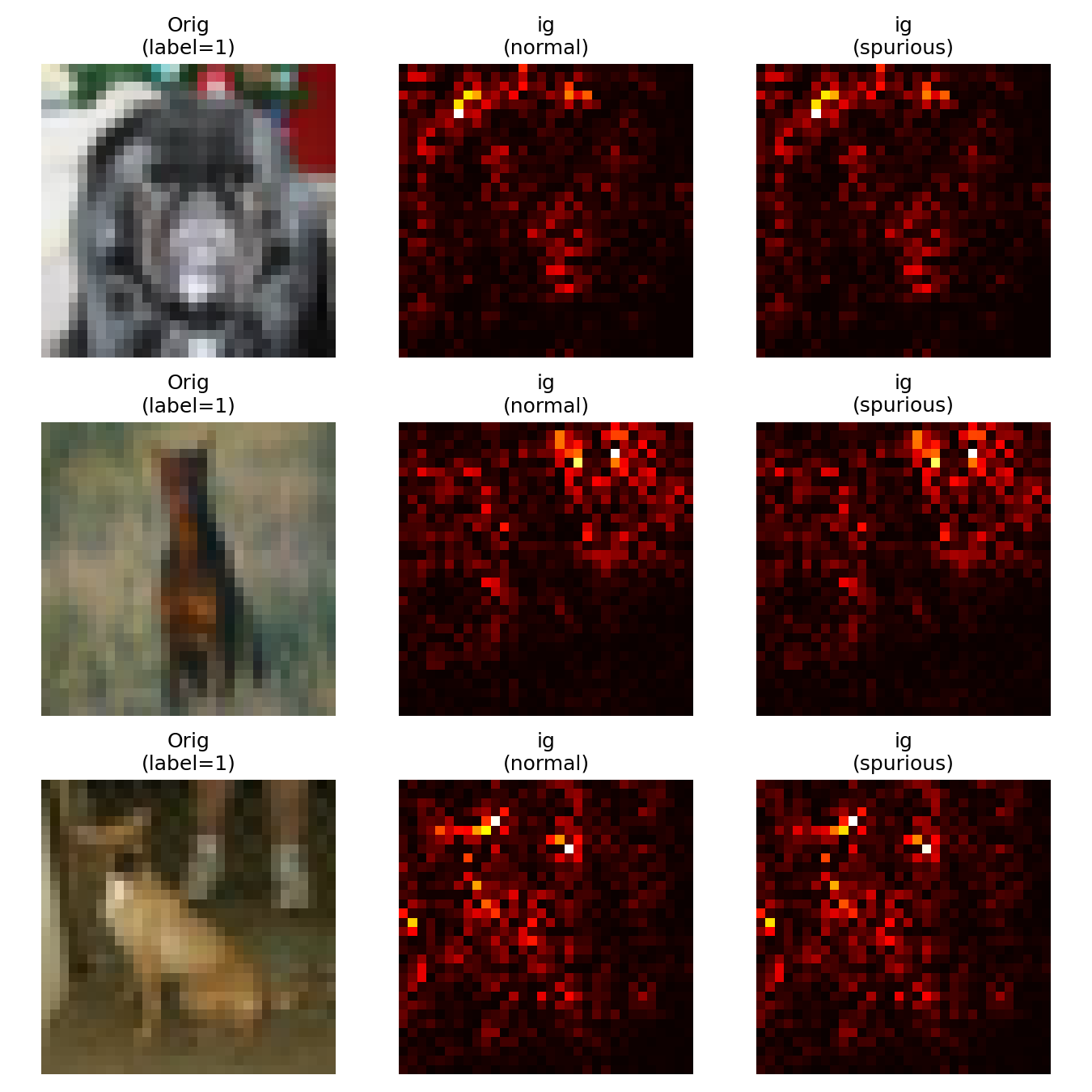}
        \caption{Case Studies on CIFAR}
\end{figure}

\begin{figure}[htpb]
    \centering
    \includegraphics[width=0.6\linewidth]{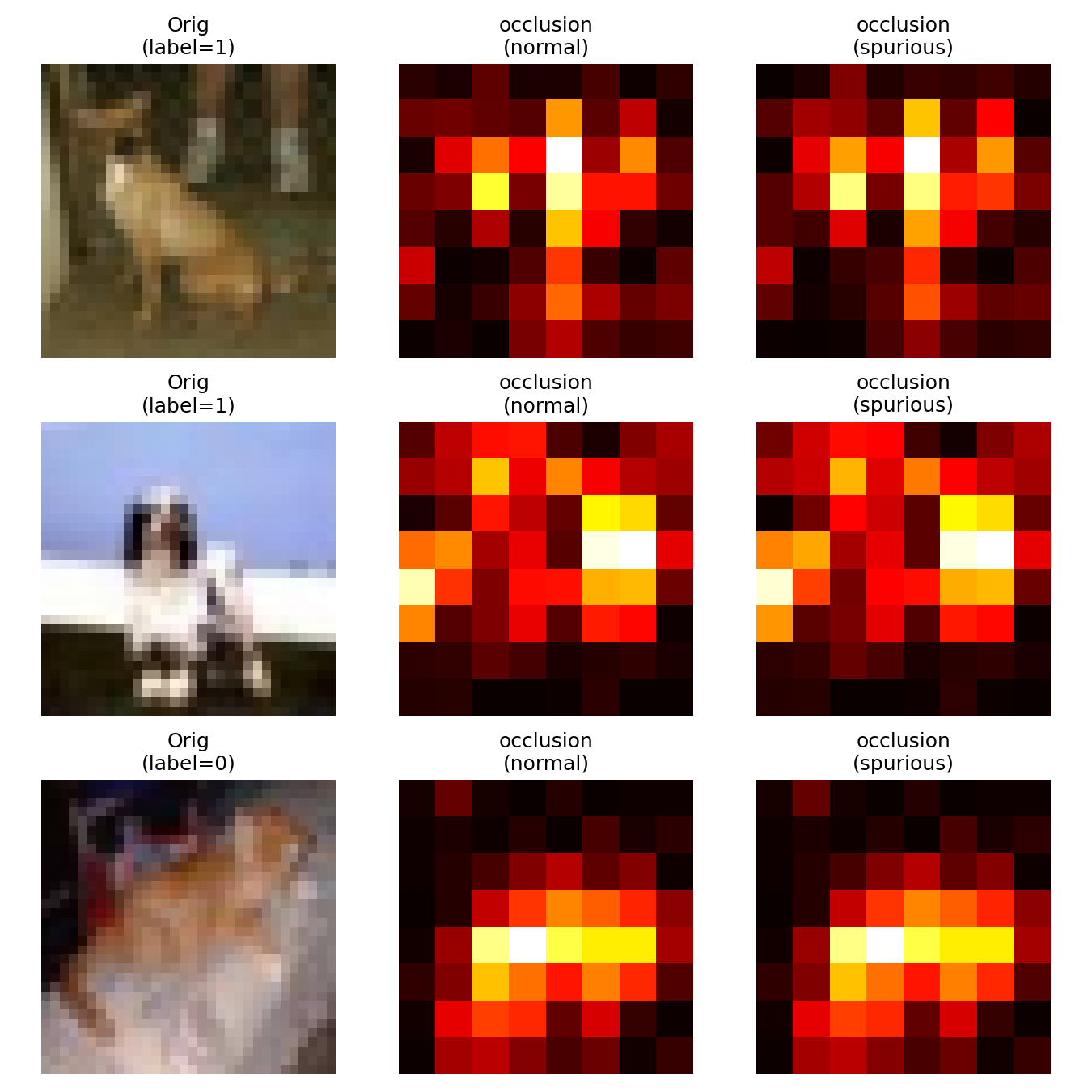}
        \caption{Case Studies on CIFAR}
\end{figure}

\end{document}